\documentclass{article}

\PassOptionsToPackage{numbers, compress}{natbib}

\usepackage[main, final]{neurips_2026}

\usepackage[utf8]{inputenc} 
\usepackage[T1]{fontenc}    
\usepackage{hyperref}       
\usepackage{url}            
\usepackage{booktabs}       
\usepackage{amsfonts}       
\usepackage{nicefrac}       
\usepackage{microtype}      
\usepackage{xcolor}         
\usepackage[most]{tcolorbox}
\usepackage{graphicx}
\usepackage{amsmath}
\usepackage{multicol}
\usepackage{multirow}
\usepackage{caption}
\DeclareMathOperator*{\argmax}{argmax}

\usepackage[table]{xcolor}
\definecolor{greyG}{rgb}{0.94, 0.88, 0.88}  
\definecolor{greyR}{rgb}{0.88, 0.94, 0.88}
\definecolor{greyB}{rgb}{0.96, 0.96, 0.85}

\title{ReCoG: Relational Concept Graph \\ for Training-Free Personalization}

\author{%
  Changseon Yu \\
  GSAIM\\
  Chung-Ang University\\
  \texttt{ckdtjs358@cau.ac.kr} \\
  \And
  Hyeokjun Kweon \\
  GSAIM\\
  Chung-Ang University\\
  \texttt{hyeokjunkweon@cau.ac.kr} \\
}

\begin{document}

\maketitle

\begin{abstract}
Training-free personalization adapts a frozen vision-language model (VLM) to recognize user-defined concepts from only a few exemplars by retrieving concept entries at inference time. 
While recent retriever-based methods enrich each concept with descriptive evidence, they largely treat concepts in isolation, so the retrieved information can be plausible yet non-discriminative, especially when the personalized concept set contains highly confusable, fine-grained categories.
We argue that personalization is inherently \textbf{relational}: correct recognition depends on the specific distinctions between a concept and its nearest alternatives.
To this end, we propose \textbf{ReCoG}, a \textbf{Re}lational \textbf{Co}ncept \textbf{G}raph for retrieval-augmented personalization.
ReCoG represents concepts as nodes and augments them with directed edges that explicitly capture discriminative cues between concept pairs, elicited by a frozen VLM during database construction.
At inference, ReCoG retrieves a shortlist and resolves ambiguity via explicit pairwise comparisons conditioned on the corresponding relational edges, selecting the concept that consistently wins against competing candidates.
We also introduce \textbf{FINGER} benchmark to test personalized recognition on confusable concept sets.
Across FINGER, ReCoG substantially outperforms prior methods, with the largest gains in the most confusable regimes.
ReCoG also achieves consistently best performance on established personalization benchmarks including captioning and VQA, confirming that relational evidence is broadly beneficial beyond fine-grained settings.
Project Page: \url{https://csy-98.github.io/ReCoG/}
\end{abstract}

\section{Introduction}
\label{sec:intro}

Personalizing vision-language model (VLM) to user-specific concepts is a central capability for interactive AI~\cite{gao2023retrieval,lewis2020retrieval,alaluf2024myvlm,nguyen2024yo,das2025training,hao2025rap,yeh2023meta, cohen2022my, richardson2023integrating,pi2024personalized}.
In many applications, users want a model to recognize or generate content consistent with a set of personal concepts, such as similar-looking products, given only a few examples.
While fine-tuning can achieve high fidelity, it is often costly, slow, and difficult to repeat whenever new concepts are introduced.

This has motivated a line of training-free personalization methods~\cite{hao2025rap,das2025training,shi2024instantbooth, kim2025driftdecodingtimepersonalizedalignments,rout2024rbmodulationtrainingfreepersonalizationdiffusion, lee2025trainingfreestylepersonalizationsvdbasedfeature,kim2025mmpbitstimemultimodal}.
This approach first builds a lightweight concept database, which comprises the sample images and features of the personal concepts.
Here, the features are typically visual embeddings or text descriptions obtained by VLM, by using the sample images as inputs.
At the time of inference, when a query is given, the relevant concepts are retrieved from the established concept database.
The retrieved concepts are used as input to the frozen backbone VLM.
Recent retriever-based approaches such as RAP~\cite{hao2025rap} and R2P~\cite{das2025training} have shown that this paradigm can be effective and scalable.

\begin{figure}[t]
  \centering
  \includegraphics[width=0.9\linewidth]{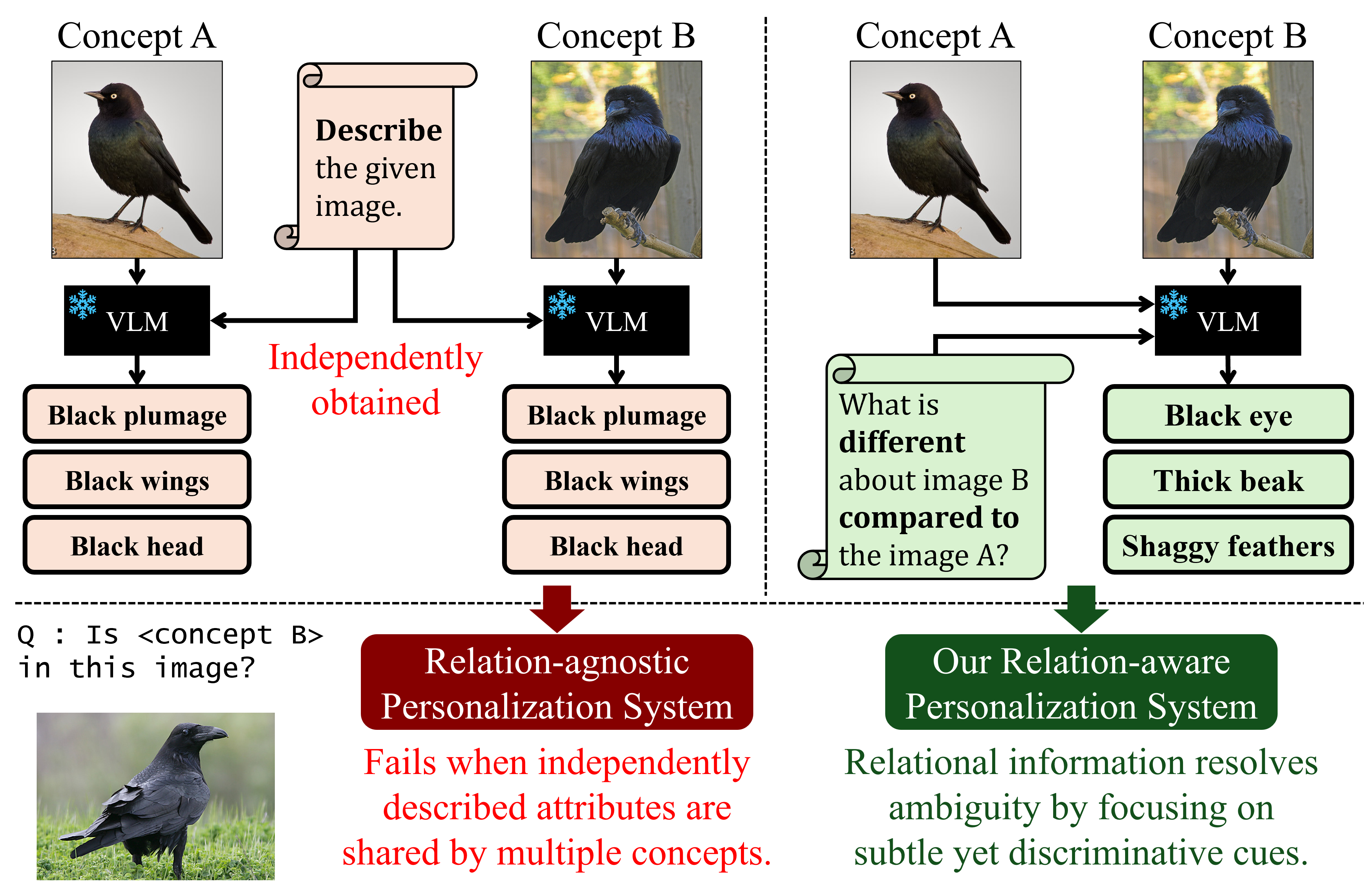}
  \vspace{-10pt}
  \caption{Illustration of why relations matter. Left: When VLM describes each concept independently, it produces salient yet generic attributes shared across confusable concepts, leading to ambiguity. Right: When VLM is conditioned on the inter-concept relation (\textit{e.g.}, asking what differs between candidates), it can elicit discriminative cues that ease recognition even under fine-grained setting.}
  \vspace{-10pt}
  \label{fig:intro}
\end{figure}

However, we find that such approaches focus on representations that primarily describe each concept in isolation, \textit{i.e.}, without considering the other concepts.
Even if plausible descriptions exist for each concept, they are not guaranteed to be helpful for identifying that concept with respect to other concepts in the personalized set.
As depicted in Fig.~\ref{fig:intro}, a description may capture the most salient properties of a concept, but still misses the less salient yet important cues for recognition.
These challenges become particularly severe in fine-grained regimes where concepts are highly similar.
This suggests that concept identity in personalization should be formulated relationally rather than purely unary.
What matters is not only a description, but also the \textbf{distinction} that makes discrimination.

In this light, we propose to consider \textbf{relations among concepts} for personalization more explicitly.
Training-free personalization can become more discriminative and more robust, by eliciting contrastive evidence during database construction and by performing relative comparisons among competing candidates during retrieval.
To this end, we propose \textbf{\underline{Re}}lational \textbf{\underline{Co}}ncept \textbf{\underline{G}}raph (\textbf{ReCoG}), a novel training-free personalization method based on the relations among the concepts, by explicitly encoding and exploiting them.
ReCoG represents concepts as nodes and introduces directed edges that store discriminative textual information specific to a concept pair.
These edges capture the attributes and cues that are most useful for distinguishing two nearby concepts, instead of repeating generic descriptions that may be shared by multiple nodes.

At inference time, we shift the focus from retrieving the most similar node to retrieving the most helpful distinction for disambiguation.
ReCoG performs pairwise comparisons between candidates using the relational edges associated with each pair, identifying which concept better matches the query under the relevant pair-specific distinctions.
This relational structure allows the system to select evidence that is aligned with the actual ambiguities present in the query.

Furthermore, we introduce a FINe-Grained pERsonalization (\textbf{FINGER}) benchmark to evaluate fine-grained personalization.
First, we employ CUB-200-2011~\cite{CUB}, a classic fine-grained bird dataset, and construct two regimes: CUB-Easy, formed by randomly sampling species, and CUB-Hard, formed by selecting highly similar species using CUB’s attribute annotations to intentionally maximize inter-concept confusability. 
Second, we include a CLEVR-based task~\cite{johnson2017clevr} where each concept corresponds to a scene configuration and images rendered from different viewpoints belong to the same concept, providing a controlled environment with systematically adjustable difficulty. 

On FINGER benchmark, ReCoG remarkably outperforms prior approaches, demonstrating that explicitly encoding distinctions is crucial for overcoming fine-grained confusion where relation-agnostic descriptions often fail. 
In addition, on conventional benchmarks~\cite{alaluf2024myvlm,nguyen2024yo,das2025training}, ReCoG also consistently achieves best performance on all tasks (recognition, image captioning, and VQA).
This strongly confirms that ReCoG is a promising framework even under coarse-grained settings.

\section{Related Works}
\label{sec:rw}

\subsection{Personalization on VLMs and MLLMs}
Recent vision-language models (VLMs) and multimodal large language models (MLLMs) have shown strong capabilities in various multimodal tasks~\cite{yao2024minicpm,jia2022visual,alayrac2022flamingo,bai2025qwen3,grattafiori2024llama}. 
However, their knowledge is mostly generic.
They can recognize common categories such as ``dog'', but they do not naturally recognize user-specific instances such as ``user's own dog". 
Personalization of VLMs/MLLMs aims to incorporate user-provided concepts into downstream responses.
Existing work has studied how frozen VLM representations can be extended to such concepts through learned embeddings, concept-specific modules, or lightweight alignment mechanisms~\cite{cohen2022my,alaluf2024myvlm,nguyen2024yo,pi2024personalized,pham2024personalized}. 
Recent work further expands this direction from single-concept image personalization to multi-concept or online settings~\cite{an2024mc,an2025unictokens,yang2025small,oh2026contextualized,bai2025online} and video applications~\cite{shi2025pvchat,zheng2026pearl}. 
Despite their effectiveness, many of these approaches require additional optimization or auxiliary modules, which can limit scalability.

\subsection{Training-Free Personalization}
Training-free personalization avoids updating model parameters for each new user or concept. 
RAP~\cite{hao2025rap} proposes a retrieval-augmented framework for MLLM personalization based on retrieval-augmented generation (RAG)~\cite{lewis2020retrieval,gao2023retrieval}. 
It stores user-specific concepts as a form of database, retrieves relevant concepts with a multimodal retriever, and injects the retrieved concept information into the MLLM input to generate personalized responses. 
R2P~\cite{das2025training} further extracts descriptive texts for personalized concepts and compares retrieved candidate concepts via Chain-of-Thought (CoT) reasoning~\cite{wei2022chain}. 
Recent training-free systems similarly build external or lightweight personal memories by storing descriptive textual metadata, visual prompts, or concept-representative visual tokens~\cite{seifi2025personalization,xu2025jarvis,seifi2026ego}. 
However, these methods primarily represent concepts independently using descriptive evidence. 
In contrast, ReCoG focuses on relations among personalized concepts, enabling VLM to perform more effective reasoning and discrimination of user-specific concepts.

\subsection{Personalization Benchmarks}
Early benchmarks focus on whether models can learn and use user-specific concepts for personalized tasks~\cite{cohen2022my,alaluf2024myvlm,nguyen2024yo,pi2024personalized}. 
Subsequent works extend it to compositional settings, including chimeric concept-context composition and multi-concept interactions~\cite{rosasco2024concon,das2025training,an2024mc}. 
More recent benchmarks further improve realism and scale by covering diverse modalities and concept types~\cite{seifi2025personalization,bai2025online,an2025unictokens}. 
Despite these contributions, benchmarks with carefully controlled fine-grained distinctions among personalized concepts remain limited. 
For example, Yo'LLaVA++ introduced in Jarvis~\cite{xu2025jarvis} targets fine-grained recognition, but it is limited to text modality. 
Since distinguishing subtle differences among user-specific visual concepts is crucial for VLM personalization, we introduce FINGER, a novel benchmark designed to evaluate fine-grained personalization across varying levels of difficulty.

\section{Preliminaries}
\label{sec:pre}

\subsection{Problem Formulation}
\label{sec:pf}
In this paper, we study a training-free personalization setting where a pre-trained VLM is adapted to recognize a user-defined set of concepts without updating its parameters. 
To enable a frozen VLM to handle novel concepts, we follow a two-stage paradigm commonly adopted in prior works: 

\noindent \: \textbf{(i)} Build a lightweight database from a small support set per concept.
\vspace{-3pt}

\noindent \: \textbf{(ii)} Retrieve relevant entries from the database to guide the VLM.

Let $\mathcal{C}=\{1,\dots,N\}$ denotes the indices of $N$ novel concepts $\{c_i\}_{i=1}^{N}$ for personalization.
We assume that each concept is associated with a small support set of exemplars:
\begin{equation}
\mathcal{S}_i = \{(x^{s}_{i,m},z^{s}_{i,m})\}_{m=1}^{M_i},
\end{equation}
where $x^{s}_{i,m}$ is the $m$-th support image for the $i$-th concept.
Notably, $z^{s}_{i,m}$ is optional information associated with $x^{s}_{i,m}$, such as textual description or embedding vectors.
For simplicity, we focus on the single-exemplar case, \textit{i.e.}, the number of exemplars $M_i=1$ and omit the exemplar index $m$.
The full personalization database is constructed by combining the entire support sets: $\mathcal{S}=\bigcup_{i=1}^{N}\mathcal{S}_i$.

This database serves as a source of personal knowledge at inference time, where the retriever selects the concept relevant to an input query.
Specifically, given a query image $x^q$ with an unknown concept label $y\in\mathcal{C}$, the goal is to predict the correct concept.
To this end, $\texttt{Retriever}$ returns the concept that is most relevant to the query from the database $\mathcal{S}$:
\begin{equation}
\hat{y} = \texttt{Retriever}(x^q,\mathcal{S}),
\end{equation}
where $\hat{y}\in\mathcal{C}$ denotes the top-1 retrieved concept.

\subsection{Conventional Relation-Agnostic Approaches}

To the best of our knowledge, existing retrieval-based personalization methods construct the database $\mathcal{S}$ without considering relations among concepts.
We refer to these approaches as ``Relation-Agnostic''.
In such methods, each concept is represented independently by a standalone entry.
We review representative relation-agnostic instantiations under the formulation in Sec.~\ref{sec:pf}.

A widely used strategy is to employ a pre-trained CLIP model.
Let $\texttt{CLIP}_{v}$ denote the CLIP visual encoder.
For each concept $c_i$, we compute and store the CLIP embedding of its support image:
\begin{equation}
f^{s}_{i} = \texttt{CLIP}_{v}(x^{s}_i) \quad \text{and} \quad z^{s}_{i} \leftarrow \{f^{s}_{i}\}.
\label{eq:clip}
\end{equation}

At inference time, the stored embeddings are used for retrieval.
Specifically, given a query image $x^q$, we compute its embedding $f^{q}=\texttt{CLIP}_{v}(x^{q})$ and perform nearest-neighbor search over the database:
\begin{equation}
\hat{y} = \texttt{Retriever}(x^q,\mathcal{S})
= \argmax_{i\in\mathcal{C}} \ \mathrm{sim}\!\left(f^{q}, f^{s}_i\right),
\end{equation}
where $\mathrm{sim}(\cdot,\cdot)$ can be a cosine similarity or a negative Euclidean distance.

This approach pre-computes an embedding for each concept during database construction, so inference reduces to a simple nearest-neighbor retrieval, making it computationally efficient.
However, the representational capacity of a single CLIP feature can be limited for fine-grained personalization.
Consequently, many works have sought to enrich each concept with more informative descriptions.

For example, R2P~\cite{das2025training} further augments each concept with textual information extracted by a VLM, in addition to the CLIP embedding.
Given the support image $x^{s}_i$ for concept $c_i$, R2P employs a frozen VLM to generate a concise text that well describes the concept, and the database is expanded as:
\begin{equation}
t^{s}_i = \texttt{VLM}(x^{s}_i|P_{\text{desc}}) \quad \text{and} \quad z^{s}_{i} \leftarrow \{f^{s}_{i}, t^{s}_i\}.
\end{equation}
where $P_{\text{desc}}$ denotes the prompt for obtaining the descriptive text.

At inference time, R2P follows a two-stage pipeline.
First, it retrieves a candidate set $\mathcal{K}\subset\mathcal{C}$ by selecting the top-$K$ nearest concepts under CLIP similarity.
Second, R2P re-ranks the candidates by asking a frozen VLM to verify whether the query image $x^{q}$ belongs to each candidate, conditioned on both the support image $x^{s}_i$ and the textual descriptions $t^{s}_i$:
\begin{equation}
(\ell_i^{\mathrm{Yes}}, \ell_i^{\mathrm{No}})
= \texttt{VLM}(x^{q}, x^{s}_i, t^{s}_i | P_{\text{belong}}),
\end{equation}
where $P_{\text{belong}}$ denotes the verification prompt that instructs the model to answer in a \texttt{Yes/No} format.
R2P then selects the concept with the highest confidence for \texttt{Yes} among the candidates:
\begin{equation}
\hat{y}=\texttt{Retriever}(x^q,\mathcal{S})=\argmax_{i\in\mathcal{K}}  \frac{\ell_i^{\mathrm{Yes}}}{\ell_i^{\mathrm{Yes}} + \ell_i^{\mathrm{No}}}.
\end{equation}

\section{Methods}
\label{sec:method}

\subsection{Importance of Relation Among Concepts in Personalization}

As discussed above, prior works have focused on improving the \emph{descriptiveness} of per-concept representations (by using CLIP embeddings or augmenting them with VLM-generated textual attributes).
However, being descriptive and being discriminative are not always aligned.
Even if plausible descriptions exist for each concept, they are not guaranteed to be \emph{discriminative} with respect to other concepts in the personalized set.
A description may capture most salient properties of a concept while still missing the few attributes that distinguish it from its closest alternatives.

\begin{figure}[t]
  \centering
  \includegraphics[width=0.95\linewidth]{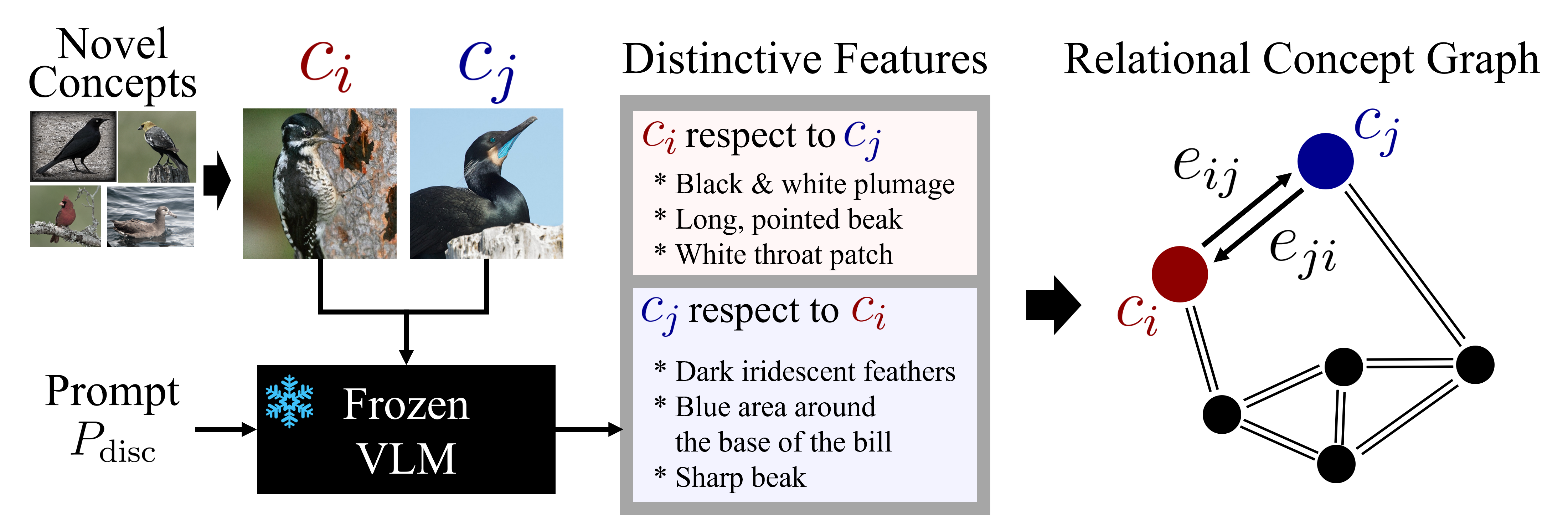}
  \vspace{-5pt}
  \caption{Given a set of novel concepts, we use a frozen VLM with a discriminative prompt $P_{\mathrm{disc}}$ to elicit directed contrastive evidence between concept pairs. For each ordered pair $(c_i,c_j)$, the VLM summarizes what makes $c_i$ different from $c_j$, producing an edge $e_{ij}$ (and vice versa). Collecting these directed edges yields the \textbf{Relational Concept Graph}, which provides explicit inter-concept relations.}
  \vspace{-5pt}
  \label{fig:db}
\end{figure}

In fine-grained regimes, \textit{i.e.}, when the concepts to be personalized are highly similar to one another, this issue becomes particularly severe.
For example, in \textbf{CUB}, many bird species share highly similar global appearances (\textit{e.g.}, ``a dark bird with a streaked breast'').
Descriptive approaches can easily focus on the \emph{most salient} attributes that characterize each species in isolation, while overlooking subtle yet crucial cues that separate visually confusable species (\textit{e.g.}, slight differences in head markings, wing-bar patterns, or bill shape).
In short, within the context of personalization, information extracted independently per concept can be descriptive yet insufficient to ensure strong separability.

A similar limitation also appears in the retrieval phase.
Most existing methods assess, for each concept independently, how likely the query belongs to that concept, and then select the concept with the highest score.
However, this strategy implicitly assumes that the resulting scores are well-calibrated and comparable across concepts.
In practice, both CLIP similarities and VLM-based verification scores can be sensitive to prompt phrasing, visual context, and concept-specific ambiguities, making their absolute scale unreliable for global ranking.
Consequently, choosing the maximum among independently computed confidences can be brittle, especially under fine-grained situation.

We argue that both limitations stem from ignoring \emph{relations among concepts}.
If database construction and retrieval are explicitly conditioned on inter-concept relations---\textit{e.g.}, by eliciting contrastive evidence during description generation and by performing relative comparisons among competing candidates during retrieval---personalization can become more discriminative and more robust.
Motivated by this insight, we propose Relational Concept Graph (\textbf{ReCoG}), which incorporates concept relations as an important signal throughout both database construction and retrieval.

\subsection{Constructing Relation-Aware Database}
\label{sec:db}

For each concept $c_i$, we compute the CLIP embedding $f^s_i$ of its support image $x^s_i$ and store it, as in Eq.~(\ref{eq:clip}).
The key difference of ReCoG is that, in addition to per-concept information, we explicitly store \textbf{relations among concepts}.
To this end, we define a directed concept graph: $\mathcal{G}=(\mathcal{C},\mathcal{E})$, where each node corresponds to a concept and edges encode their discriminative relation.

For a pair of concepts $(c_i,c_j)$, we ask a VLM to describe what makes $c_i$ different from $c_j$, as shown in Fig.~\ref{fig:db}.
We denote the resulting textual evidence as a directed edge $e_{ij}$:
\begin{equation}
e_{ij} = \texttt{VLM}(x^{s}_i, x^{s}_j \mid P_{\mathrm{disc}}),
\label{eq:edge}
\end{equation}
where $P_{\mathrm{disc}}$ is a prompt for eliciting discriminative evidence (Sec.~\ref{sec:app_prompt}).
Note that $e_{ij}$ and $e_{ji}$ are generally different, as they are produced from opposite perspectives: $e_{ij}$ describes the difference from the viewpoint of $c_i$, whereas $e_{ji}$ describes it from the viewpoint of $c_j$.
In practice, we obtain both edges at once, from a single VLM comparison for each unordered concept pair.

We then augment each concept entry with its outgoing discriminative edges:
\begin{equation}
\mathcal{S}_i = \{(x^{s}_{i},z^{s}_{i})\} \quad \text{where} \quad z^{s}_{i} \leftarrow \Big\{f^{s}_{i},\ \{e_{ij}\}_{j\in\mathcal{A}(i)}\Big\},
\label{eq:outgoing}
\end{equation}
where $\mathcal{A}(i)=\mathcal{C}\setminus\{i\}$ is the set of concepts for which we store outgoing edges.
By explicitly storing how concepts differ from one another, the database captures evidence that is directly useful for discrimination, rather than describing each concept in isolation.

\subsection{Discriminative Retrieval via Explicit Comparison}
\label{sec:retrieval}

At inference time, we begin by pruning the search space from $\mathcal{C}$ to a candidate set $\mathcal{K}\subset\mathcal{C}$ using CLIP similarity, similar to R2P.
Within $\mathcal{K}$, the candidates are close in the CLIP space, so distinguishing them requires more discriminative evidence.
Here, we perform more fine-grained retrieval using the constructed graph-based database.
Specifically, ReCoG directly performs \textbf{relative comparisons} between candidates, instead of independently scoring each candidate.

For each unordered pair $(u,v)$ in $\mathcal{K}$, we ask a frozen VLM to decide whether the query is closer to $c_u$ or $c_v$, conditioned on the relational edges $e_{uv}$ and $e_{vu}$:
\begin{equation}
(\ell_{u\succ v},\ell_{v\succ u})
= \texttt{VLM}(x^{q}, x^{s}_u, x^{s}_v, e_{uv}, e_{vu} \mid P_{\mathrm{comp}}),
\end{equation}
where $P_{\mathrm{comp}}$ is a prompt asking the model to perform a pairwise comparison (Sec.~\ref{sec:app_prompt}).
Here, $\ell_{u\succ v}$ denotes the logit indicating that the query is closer to $c_u$ than $c_v$, and vice versa.
We use both $e_{uv}$ and $e_{vu}$ since both perspectives provide complementary evidence.
The winner of this comparison is
\begin{equation}
w(u,v)=
\begin{cases}
u, & \ell_{u\succ v} > \ell_{v\succ u},\\
v, & \text{otherwise}.
\end{cases}
\end{equation}
We then aggregate match outcomes in a round-robin manner:
\begin{equation}
W_u=\sum_{v\in\mathcal{K}\setminus\{u\}} \textbf{1}\big[w(u,v)=u\big],
\end{equation}
where $\textbf{1}[\cdot]$ denotes the indicator function.
Ties are possible in principle; in such cases, we select the one with the highest CLIP similarity. Otherwise, the concept with the largest number of wins is returned:
\begin{equation}
\hat{y}=\texttt{Retriever}(x^q,\mathcal{S})=\argmax_{u\in\mathcal{K}}\ W_u.
\end{equation}

\subsection{Sparse ReCoG for Computational Efficiency}
\label{sec:sparse_graph}

Constructing relational edges for all concept pairs requires $\mathcal{O}(N^2)$ VLM calls during database construction.
Although this cost is incurred only once (offline), we additionally consider a simple sparse variant to reduce this overhead.
Instead of using the full adjacency $\mathcal{A}(i)=\mathcal{C}\setminus\{i\}$, we construct $\mathcal{A}(i)$ using CLIP similarity among support images.
Specifically, for each concept $c_i$, we select its top-$k$ nearest concepts as relational neighbors:
\begin{equation}
\mathcal{A}(i) =
\operatorname{TopK}_{j \in \mathcal{C}\setminus\{i\}}
\mathrm{sim}(f_i^s, f_j^s).
\end{equation}
We then construct relational edges only for concept pairs that appear in at least one selected adjacency set.
This yields at most $Nk$ pair proposals and often fewer actual VLM calls (due to duplication), while focusing on visually similar concepts where discriminative evidence is most likely to be useful.

At inference time, sparse construction may leave some candidate pairs without precomputed relational edges.
There are two possible ways to handle such missing edges.
The first option is to generate the missing edge on demand using the same discriminative prompt as in database construction.
This preserves the full pairwise comparison behavior for the encountered pair, at the cost of an additional inference-time VLM call; once generated, the edge can also be cached into the database for future use.
The second option is to avoid any additional VLM call and treat the missing comparison as a tie, assigning no win to either candidate.
This may introduce a small performance drop because some pairwise comparisons are skipped during round-robin, but it keeps inference strictly within the precomputed sparse graph.
In our experiments (Sec.~\ref{sec:exp_sparse}), we find that this simple no-call strategy already provides a favorable trade-off between computational cost and recognition accuracy.

\section{FINGER Benchmark}
\label{sec:benchmark}

\begin{figure}[t]
  \centering
  \includegraphics[width=0.9\linewidth]{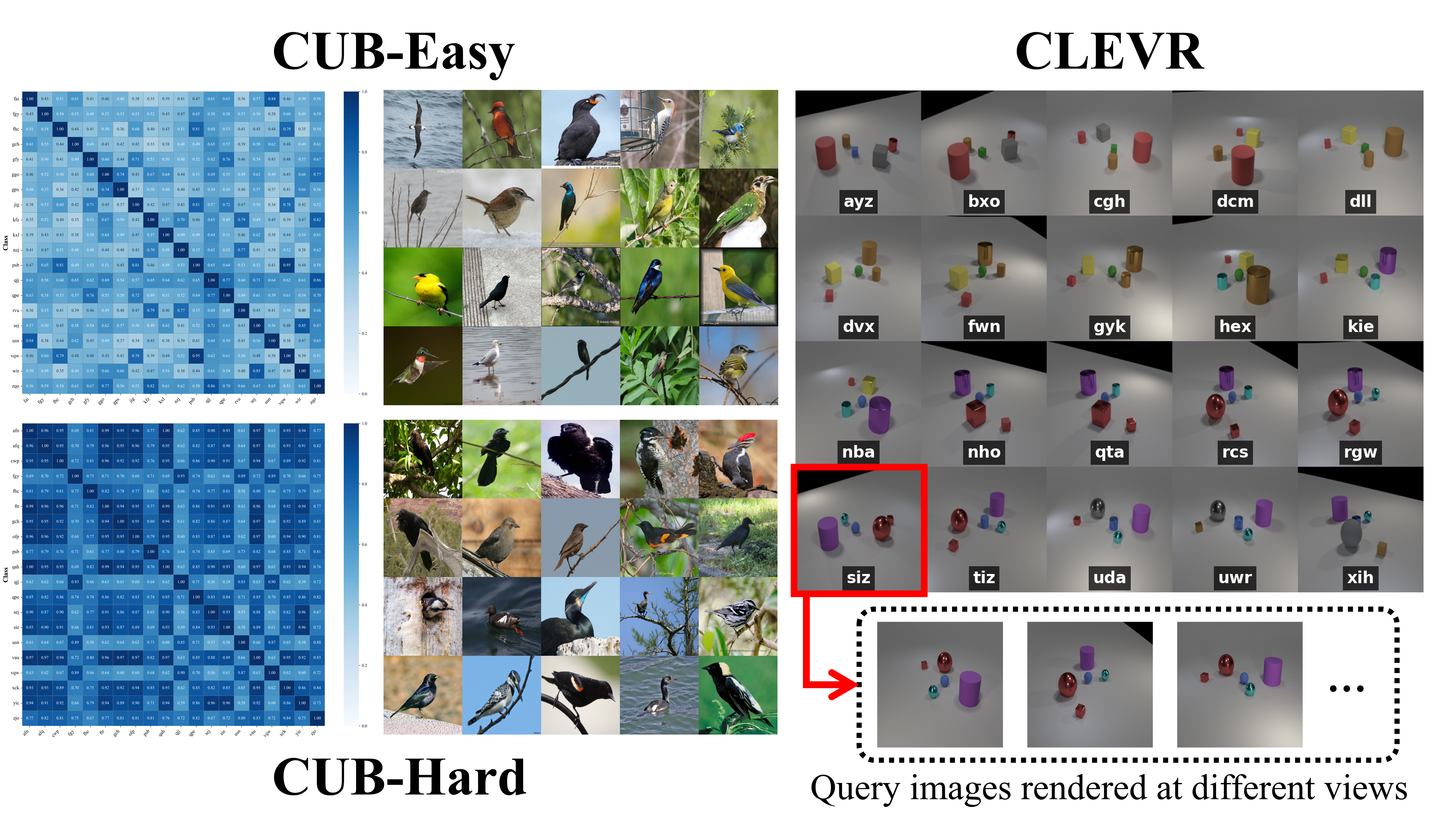}
  \vspace{-10pt}
  \caption{The proposed FINGER Benchmark. \textbf{Left}: 20 concepts in CUB-Easy (top) and CUB-Hard (bottom). We also provide the pairwise concept similarity matrices computed from CUB’s visual attribute vectors (darker means higher similarity, \textit{i.e.}, more difficult). \textbf{Right}: 20 concepts in CLEVR, where each concept corresponds to a specific scene configuration.}
  \label{fig:benchmark}
  \vspace{-10pt}
\end{figure}

In practical personalization scenarios, the challenge is often fine-grained: users may want an AI system to distinguish between highly similar concepts that differ only in subtle visual details (\textit{e.g.}, closely related species or visually similar parts in manufacturing).
Motivated by this, we introduce a \textbf{FINe-Grained pERsonalization (FINGER)} Benchmark that explicitly targets concept sets with high visual confusability, as in Fig.~\ref{fig:benchmark}.

The first task is based on CUB-200-2011 dataset~\cite{CUB}, which contains 200 bird species with subtle inter-class variations.
A useful property of CUB is that each image has a vector of semantic visual attributes (\textit{e.g.}, body color), enabling controlled construction of challenging concept sets.
Using CUB, we build two tasks with different levels of difficulty.
Each task contains $N=20$ concepts; for each concept, we provide one support image and nineteen query images, resulting in 400 images per task.

We construct \textbf{CUB-Easy} task by sampling 20 species at random from the 200 classes.
We therefore refer to this task as “Easy,” although it remains challenging.
In contrast, \textbf{CUB-Hard} task is intentionally designed to be more confusable.
Specifically, we select visually similar concepts by leveraging CUB attributes, for example by focusing on species that score highly along attributes related to \texttt{black} appearance.
CUB-Hard matches CUB-Easy in configuration, but is considerably more difficult: the system must retrieve the correct concept from a set of visually similar, predominantly black birds given a query that is itself a black bird.


We also introduce a \textbf{CLEVR}-based task to provide a more controlled environment~\cite{johnson2017clevr}.
CLEVR framework supports compositional scene generation with objects of different shapes, colors, sizes, and materials.
We treat an entire scene configuration (\textit{i.e.}, a set of objects and their spatial arrangement) as a concept, and regard images rendered from different viewpoints as belonging to that concept.
For this task, we use 20 concepts with 1 support image and 9 query images per concept (200 images total).
This benchmark is particularly useful because its difficulty can be systematically adjusted by controlling the degree of overlap in scene attributes.
More details are provided in Appendix (Sec.~\ref{sec:app_clevr}).


\section{Experimental Results}
\label{sec:exp}
\vspace{-5pt}
\subsection{Setup}
\vspace{-5pt}
\noindent\textbf{Datasets/Metrics.}
In addition to our FINGER benchmark, we evaluate ReCoG on three established benchmarks: MyVLM (29 concepts)~\cite{alaluf2024myvlm}, Yo’LLaVA (40 concepts)~\cite{nguyen2024yo}, and PerVA (329 concepts)~\cite{das2025training}.
For recognition task, we report positive accuracy (P.Acc.), measured on positive queries that include the target concept, and negative accuracy (N.Acc.), measured on negative queries that exclude the target concept. As same as R2P~\cite{das2025training}, for each target concept, we exhaustively evaluates images from all other concepts as negative examples. However, we measured the negative accuracy of the Baseline and ReCoG from their top-1 retrieval results, as the goal was to evaluate the output of the personalization retriever. Wtd. denotes the uniformly weighted average of them: (P.Acc.+N.Acc.)/2.
For captioning and VQA tasks, we report hard recall and accuracy, respectively, following R2P~\cite{das2025training}.

\noindent\textbf{Implementation.}
We construct the database using CLIP~\cite{radford2021learning} ViT-L. 
We use MiniCPM-o-2.6~\cite{yao2024minicpm} as VLM by default.
Results with LLaVA~\cite{liu2023visual} are in Appendix (Sec.~\ref{sec:app_impact_vlm}).
Similar to R2P~\cite{das2025training}, we construct database using a single reference image per concept, without negative samples. 
Although this setting is more challenging, it better matches real-world personalization with minimal user effort.

\subsection{Results on FINGER Benchmark}
Table~\ref{tab:fgpb} reports the results on our FINGER Benchmark, and it cleanly illustrates why relational understanding is the key component in training-free personalization.
\textbf{ReCoG dominates across all tasks and metrics}. 
On Wtd., ReCoG achieves 98.1\% on CUB-Easy, 92.8\% on CUB-Hard, and 93.3\% on CLEVR, outperforming the strongest prior work R2P~\cite{das2025training} by +9.7, +11.8, and +13.7 percent points, respectively. 
The gains are even larger on P.Acc. 
This is notable because our benchmark comprising highly confusable concepts, yet ReCoG still maintains substantial performance.

These results also highlight how the conventional relation-agnostic personalization approach struggles with fine-grained confusion.
R2P~\cite{das2025training} is a substantially strong method that performs well on conventional benchmarks (refer to Table~\ref{tab:conventional_bench}).
On FINGER benchmark, however, it faces difficulty due to the visually confusing concepts (\textit{e.g.}, 63.8\% P.Acc. on CUB-Hard and 61.1\% P.Acc. on CLEVR). 
This behavior is consistent with our motivation: concept descriptions generated in isolation are often plausible but fail to identify the specific distinctions needed to separate a concept from its closest confounders.

Finally, the comparison between R2P~\cite{das2025training} and Baseline isolates the role of retrieval strategy. Our Baseline uses the same descriptive features as R2P~\cite{das2025training} but replaces separated verification with comparison-based retrieval. 
If pairwise retrieval alone were the main source of gain, we would expect a meaningful improvement over R2P~\cite{das2025training} instead, Baseline is worse. 
This implies that pairwise comparison is not always beneficial when the evidence is not aligned with the real ambiguity. 

Fig.~\ref{fig:conf} shows ReCoG's superiority intuitively. The left confusion matrix corresponds to Baseline, and the right matrix corresponds to ReCoG.
The baseline exhibits substantial off-diagonal mass. This pattern is consistent with the nature of CUB-Hard, where concepts share very similar global appearance.
In contrast, ReCoG produces a near-diagonal structure, indicating that ReCoG consistently reduces confusion across the entire fine-grained concept set.
More qualitative results of ReCoG on FINGER benchmark can be found in Appendix (Sec.~\ref{sec:app_qual}).

Table~\ref{tab:hop_main} further provides a controlled analysis of CLEVR difficulty.
Here, $K$ denotes scene complexity and hop represets distance between the concepts (i.e., 1-hop is more fine-grained).
For more details, please refer to Appendix (Sec.~\ref{sec:app_clevr}).
As $K$ increases, all methods generally suffer a drop in positive accuracy, indicating that larger scenes introduce more distractors and make personalized matching harder.
Nevertheless, ReCoG consistently outperforms R2P and the Baseline.
The margin is especially pronounced in the more confusable 1-hop regime, where concepts differ only by a small local change.
Even in the relatively easier 2-hop setting, ReCoG maintains a clear advantage, reaching 81.4\% compared with 68.4\% for R2P~\cite{das2025training} and 40.0\% for the Baseline.

\begin{table*}[t]
\centering
\caption{Recognition and captioning performance on the proposed FINGER benchmark. Baseline denotes the setting using the relation-agnostic description (the same as R2P~\cite{das2025training}) and performs retrieval in a pairwise manner in Sec.~\ref{sec:retrieval}. ReCoG uses the relation-aware concept graph and pairwise retrieval.}
\footnotesize
\label{tab:fgpb}
\resizebox{\textwidth}{!}{
\setlength{\tabcolsep}{3pt}
\begin{tabular}{l|ccc|c|ccc|c|ccc|c}
\hline
\multirow{2}{*}{Methods} & \multicolumn{4}{c|}{CUB-Easy} & \multicolumn{4}{c|}{CUB-Hard} & \multicolumn{4}{c}{CLEVR-Default} \\
\cline{2-13}
 & P.Acc & N.Acc & Wtd. & Recall & P.Acc & N.Acc & Wtd. & Recall & P.Acc & N.Acc & Wtd. & Recall \\
\hline
R2P \cite{das2025training} & 87.9 & 99.4 & 93.6 & 76.1 & 63.8 & 98.1 & 81.0 & 55.0 & 61.1 & 98.0 & 79.6 & 24.4 \\
Baseline & 50.0 & 97.3 & 73.7 & 65.0 & 41.8 & 96.9 & 69.4 & 47.9 & 51.7 & 97.5 & 74.6 & 14.8 \\
\hline
ReCoG (\textbf{Ours}) & \textbf{96.3} & \textbf{99.8} & \textbf{98.1} & \textbf{80.8} & \textbf{86.3} & \textbf{99.3} & \textbf{92.8} & \textbf{62.1} & \textbf{87.2} & \textbf{99.3} & \textbf{93.3} & \textbf{41.7} \\
\hline
\end{tabular}
}
\end{table*}
\begin{figure}[t]
\centering
\vspace{-5pt}

\begin{minipage}[t]{0.64\textwidth}
    \centering
    \vspace{0pt}
    \includegraphics[width=0.99\linewidth]{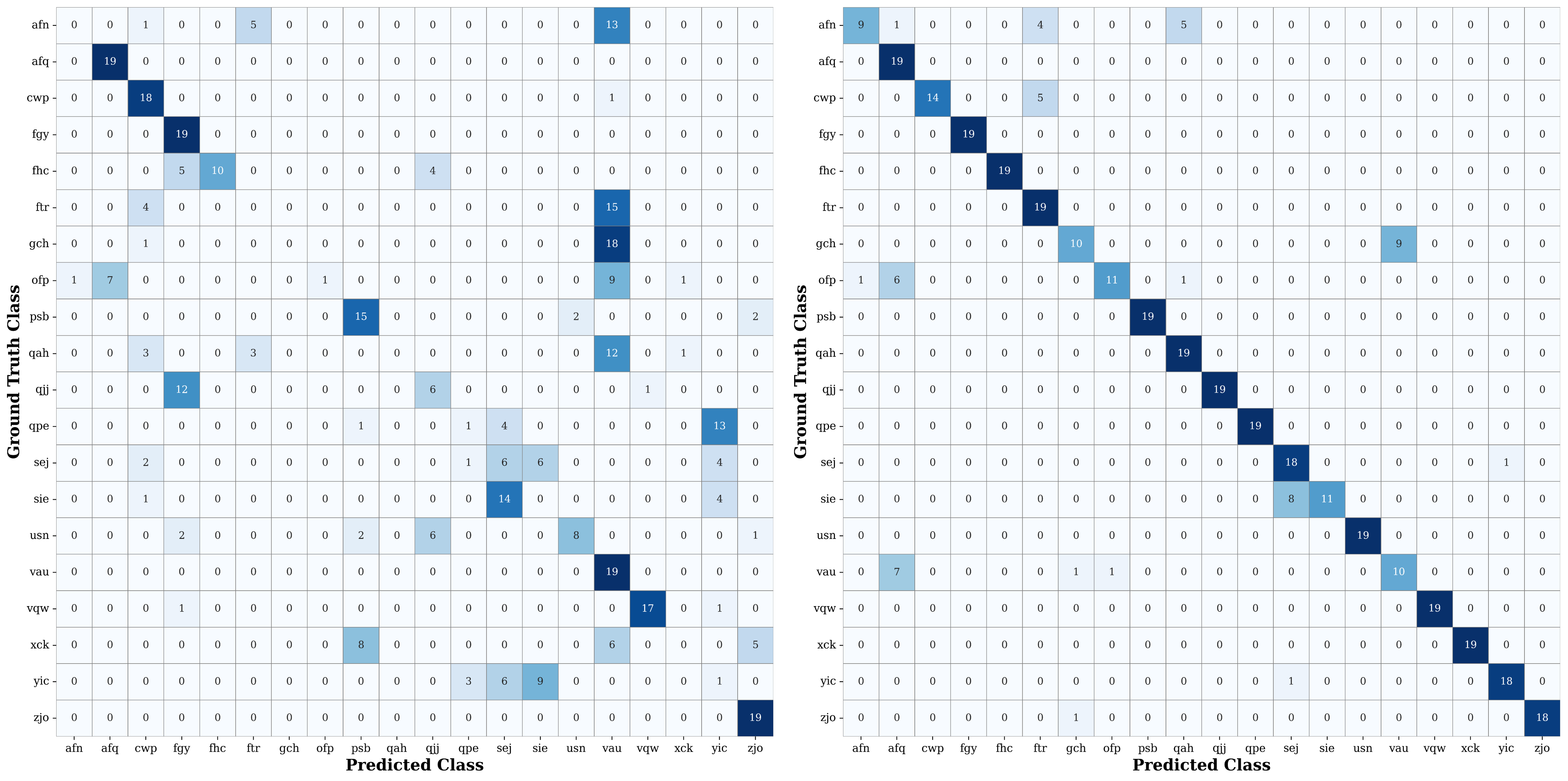}
    \vspace{-5pt}
    \captionof{figure}{Conf. matrices of Baseline (left) and ReCoG (right).}
    \label{fig:conf}
\end{minipage}
\hfill
\begin{minipage}[t]{0.34\textwidth}
    \centering
    \vspace{0pt}
    \resizebox{0.99\textwidth}{!}{%
    \setlength{\tabcolsep}{2pt}
    \begin{tabular}{c|l|c|c}
    \hline
    \textbf{K} & \textbf{Methods} & \textbf{1-hop} & \textbf{2-hop} \\
    \hline
    \multirow{3}{*}{\textbf{4}} 
      & R2P      & 61.7 / 98.0 & 75.0 / 98.7 \\
      & Baseline & 52.8 / 97.5 & 68.3 / 98.3 \\
      & ReCoG    & \textbf{91.4 / 99.5} & \textbf{91.7 / 99.6} \\
    \hline
    \multirow{3}{*}{\textbf{5}} 
      & R2P      & 53.8 / 97.6 & 72.2 / 98.5 \\
      & Baseline & 36.1 / 96.6 & 50.5 / 97.4 \\
      & ReCoG    & \textbf{76.6 / 98.8} & \textbf{82.2 / 99.1} \\
    \hline
    \multirow{3}{*}{\textbf{6}} 
      & R2P      & 46.6 / 97.2 & 68.4 / 98.3 \\
      & Baseline & 30.0 / 96.3 & 40.0 / 96.8 \\
      & ReCoG    & \textbf{74.4 / 98.7} & \textbf{81.4 / 99.0} \\
    \hline
    \end{tabular}%
    }
    \captionof{table}{P.Acc/N.Acc comparison across different $K$ and hop distances on CLEVR task.}
    \label{tab:hop_main}
\end{minipage}

\vspace{-5pt}
\end{figure}

\begin{table*}[t]
\centering
\caption{Recognition/captioning performance on~\cite{cohen2022my,nguyen2024yo,das2025training} and VQA performance on Yo'LLaVA}
\vspace{-5pt}
\begin{minipage}[t]{0.76\textwidth}
\vspace{0pt}
\centering
\resizebox{\textwidth}{!}{
\setlength{\tabcolsep}{3pt}
\begin{tabular}{l|ccc|c|ccc|c|ccc|c}
\hline
\multirow{2}{*}{Methods} & \multicolumn{4}{c|}{MyVLM\cite{alaluf2024myvlm}} & \multicolumn{4}{c|}{Yo'LLaVA\cite{nguyen2024yo}} & \multicolumn{4}{c}{PerVA\cite{das2025training}} \\
\cline{2-13}
 & P.Acc & N.Acc & Wtd. & Recall & P.Acc & N.Acc & Wtd. & Recall & P.Acc & N.Acc & Wtd. & Recall \\
\hline
MyVLM & 96.6 & 90.9 & 93.8 & 0.02 & - & - & - & - & 66.0 & 58.5 & 62.2 & - \\
Yo'LLaVA & 97.0 & 95.7 & 96.4 & 0.1 & 94.9 & 89.8 & 92.4 & 0.2 & 75.1 & 69.0 & 72.0 & - \\
\hline
RAP & 94.4 & 98.8 & 96.6 & 88.0 & 86.0 & 99.2 & 92.2 & 79.7 & 92.9 & 85.2 & 89.0 & 64.1 \\
R2P & 96.3 & 98.5 & 97.4 & 91.4 & 91.1 & 97.7 & 94.4 & 87.1 & 90.2 & 92.8 & 91.8 & 72.5 \\
\hline
ReCoG & \textbf{98.6} & \textbf{99.9} & \textbf{99.3} & \textbf{92.9} & \textbf{95.0} & \textbf{99.9} & \textbf{97.4} & \textbf{90.4} & \textbf{97.7} & \textbf{99.9} & \textbf{98.8} & \textbf{73.7} \\
\hline
\end{tabular}
}
\end{minipage}
\hfill
\begin{minipage}[t]{0.23\textwidth}
\vspace{0pt}
\centering
\resizebox{\textwidth}{!}{
\setlength{\tabcolsep}{2pt}
\begin{tabular}{lc}
\hline
Methods & Accuracy \\ \hline
Yo'LLaVA~\cite{nguyen2024yo} & 92.9 \\
RAP~\cite{hao2025rap} & 93.2 \\
GPT-4V+VP~\cite{nguyen2024yo} & 86.6 \\
LLaVA~\cite{nguyen2024yo} & 89.9 \\
R2P~\cite{das2025training} & 96.5 \\ \hline
ReCoG & \textbf{97.0} \\ \hline
\end{tabular}
}
\end{minipage}
\label{tab:conventional_bench}
\vspace{-10pt}
\end{table*}

\begin{figure}[t]
  \centering
  \includegraphics[width=0.99\linewidth]{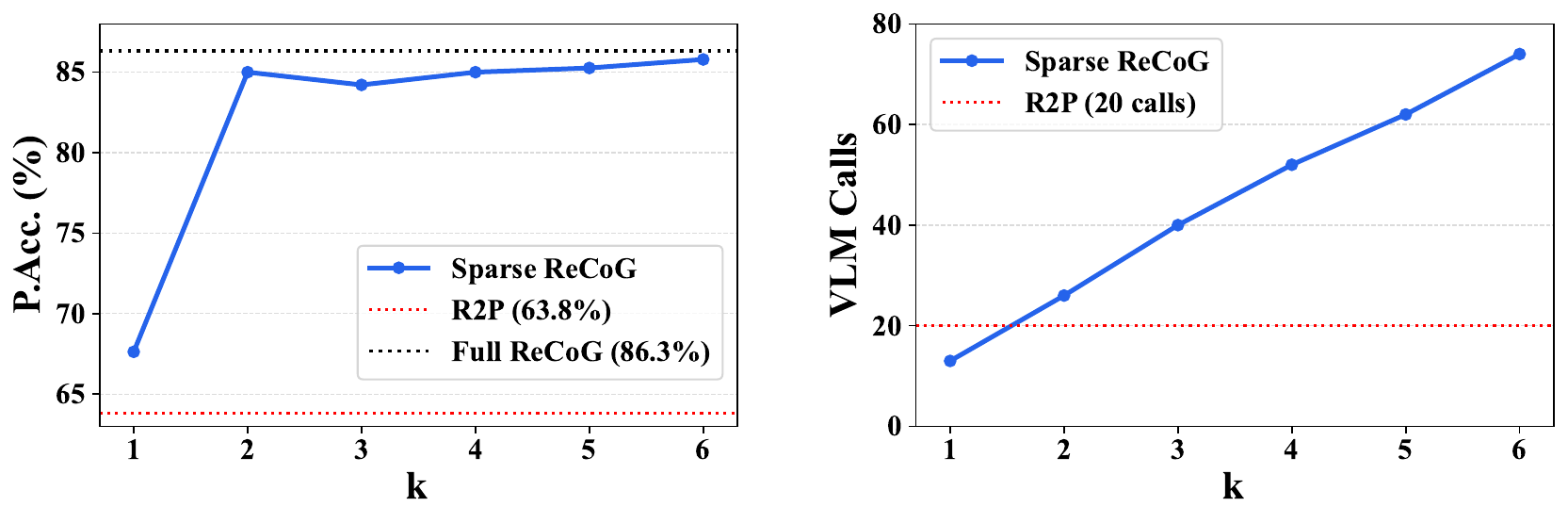}
  \vspace{-10pt}
  \caption{Effect of sparse graph construction on CUB-Hard. Left: P.Acc. of ReCoG when relational edges are sparsely constructed. Right: the number of VLM calls required for sparse database construction after removing duplicated concept pairs. Red dashed lines are those of R2P~\cite{das2025training}.}
  \vspace{-10pt}
  \label{fig:sparse}  
\end{figure}

\subsection{Results on Conventional Personalization Benchmarks}
\vspace{-7pt}
Table~\ref{tab:conventional_bench} reports results on three established benchmarks (MyVLM~\cite{alaluf2024myvlm}, Yo'LLaVA~\cite{nguyen2024yo}, and PerVA~\cite{das2025training}), comparing ReCoG against both training-based~\cite{alaluf2024myvlm,nguyen2024yo} and training-free~\cite{hao2025rap,das2025training} approaches. 
ReCoG achieves the best performance across all tasks (recognition, captioning, and VQA) and all benchmarks, consistently outperforming prior methods under the same evaluation protocol.
The results indicate that the relational evidence encoded by ReCoG is not only beneficial in our challenging FINGER benchmark, but also transfers to conventional personalization scenarios where concepts are less confusable, yielding substantial performance even on the relatively saturated benchmarks.  
\vspace{-5pt}

\subsection{Analysis on Sparse ReCoG}
\label{sec:exp_sparse}
\vspace{-7pt}

We evaluate sparse ReCoG on CUB-Hard to analyze the trade-off between database construction cost and recognition accuracy.
In this experiment, we use the tie strategy in Sec.~\ref{sec:sparse_graph}: if a candidate pair does not have precomputed relational edges, the comparison is treated as a tie during round-robin aggregation.
Alternatively, if additional inference-time VLM calls are acceptable occasionally, missing edges can be constructed online. This yields the same behavior as full ReCoG.

As shown in Fig.~\ref{fig:sparse}, increasing $k$ rapidly improves performance with only a small number of additional VLM calls.
Even with $k=1$, Sparse ReCoG already outperforms R2P~\cite{das2025training} while requiring only 13 VLM calls for graph construction.
More notably, with $k=2$, it reaches 85.0\% P.Acc. using only 26 VLM calls, which is close to the full ReCoG performance of 86.3\% obtained with all 190 pairwise edges.
After this point, the performance largely saturates, suggesting that most useful relational evidence comes from a small set of visually similar neighboring concepts.
These results indicate that sparse ReCoG could provide a practical cost-accuracy trade-off.
\vspace{-5pt}

\subsection{Isolating Relational Evidence from Pairwise Retrieval}
\label{sec:exp_sparse}
\vspace{-7pt}

Table ~\ref{tab:ablation_scoring} reports an ablation study on relational edge evidence and candidate scoring to isolate the effects of relational reasoning from the pairwise search approach. ReCoG consists of two distinct components: (i) relational evidence construction and (ii) pairwise retrieval. We therefore organize the ablation on CUB-Hard along two axes to distinguish their respective contributions: 

\begin{table}[t]
\centering
\caption{Ablation study on different evidence types and scoring procedures on CUB-Hard.}
\label{tab:ablation_scoring}
\begin{tabular}{l|c|c}
\hline
 & \textbf{Candidate-wise scoring} & \textbf{Pairwise scoring} \\
\hline
\textbf{Concept-wise descriptive evidence} & Method 1 (R2P~\cite{das2025training}): 63.8\% & Method 2: 41.8\% \\
\hline
\textbf{Relational evidence} & Method 3: 74.0\% & Method 4 (ReCoG): \textbf{86.3\%} \\
\hline
\vspace{-20pt}
\end{tabular}
\end{table}

Method 1 is R2P~\cite{das2025training}. It independently constructs a description for each concept and verifies each retrieved candidate separately. Method 4 is our ReCoG, which constructs pair-specific relational evidence and uses it during pairwise comparison and round-robin aggregation. To further isolate the contribution of each component—relational evidence and pairwise scoring—we additionally evaluate Methods 2 and 3. Method 2 is the Baseline, which uses the same concept-wise descriptions as R2P~\cite{das2025training} but replaces candidate-wise verification with the pairwise retrieval procedure used by ReCoG.
Finally, Method 3 is the requested relational-evidence variant with candidate-wise scoring. For each candidate, we aggregate the descriptions directed toward that candidate from all relational edges involving the other retrieved candidates. We then independently evaluate each candidate using the query image, its support image, and its aggregated relational evidence, following the candidate-wise verification procedure of R2P~\cite{das2025training}. The candidate with the highest “Yes” confidence is selected.

Method 3 improves P.Acc. from 63.8\% to 74.0\%, showing that contrastively generated relational evidence is more informative than independently generated concept descriptions. However, it remains 12.3 percentage points below ReCoG. These results demonstrate that contrastively generated relational evidence is an important source of improvement, but is not sufficient to explain the full performance of ReCoG. They also show that pairwise retrieval becomes substantially more effective when it is conditioned on pair-specific relational evidence.
\vspace{-5pt}

\subsection{Scalability to Large and Dynamic Concept Libraries}
\label{sec:exp_sparse}
\vspace{-7pt}

To directly analyze the scalability of ReCoG, we conducted an additional large-scale experiment using all 200 species in CUB, treating each species as a distinct personalized concept. 
For the missing edges in Sparse ReCoG, we consider two strategies. In the tie strategy, a candidate pair without a precomputed relation edge is treated as a tie. Consequently, Sparse ReCoG may perform fewer than pairwise comparisons. For example, the tie strategy requires an average of 2.46 VLM calls per query (2.79 s), compared with 3 calls for R2P~\cite{das2025training} (3.83 s).
In the on-demand strategy, a missing relation is generated when the corresponding candidate pair is encountered for the first time and is then cached for all subsequent queries. Under our default setting of , this requires at most three additional calls. However, the same pair never requires another construction call once its relation has been cached. In our 200-concept experiment, the on-demand strategy requires an average of 3.281 VLM calls per query (3.86 s) during the first evaluation pass, and this decreases to 3 calls per query after caching.
Even with and the no-call tie strategy, Sparse ReCoG (56.2\%) outperforms R2P~\cite{das2025training} (48.6\%) in P.Acc., while requiring fewer VLM calls in both offline database construction and online inference. 

\vspace{-5pt}

\section{Conclusion}
\vspace{-10pt}
We presented ReCoG, a training-free personalization framework that explicitly models relations among user-defined concepts. 
ReCoG constructs a Relational Concept Graph whose directed edges encode pair-specific discriminative evidence. 
At inference time, ReCoG leverages these relational edges to perform explicit pairwise comparisons among retrieved candidates.
We further introduced FINGER, a benchmark designed to evaluate personalized recognition under controlled fine-grained difficulty. 
Extensive experiments show that ReCoG substantially improves over prior training-free personalization methods on FINGER, especially in the most confusable regimes. 

\noindent\textbf{Limitation.}
Although sparse ReCoG shows that the construction cost can be substantially reduced while maintaining strong performance, scalability may still become challenging when the number of concepts grows significantly or when the database is updated online. 
Developing more adaptive and efficient graph sparsification strategies is therefore an interesting direction for future work.

\section*{Acknowledgements and Disclosures}
This research was supported by the ANCHOR through the Seoul ANCHOR Center, funded by the Ministry of Education (MOE) and the Seoul Metropolitan Government (2026-ANCHOR-01-024-06), and the National Research Foundation of Korea(NRF) grant funded by the Korea government(MSIT) (RS-2026-25510069).


\bibliographystyle{splncs04}
\bibliography{main}


\clearpage

\appendix

\renewcommand{\thefigure}{A\arabic{figure}}
\renewcommand{\thetable}{A\arabic{table}}

\section{Additional Details and Experiments on CLEVR}
\label{sec:app_clevr}

We construct a FINGER benchmark using a CLEVR-based pipeline.
Scene rendering is implemented with custom scripts built on top of the original CLEVR utilities, using a fixed base scene and predefined object/material dictionaries.
Each concept is defined over object-level attributes from a controlled vocabulary, including shape (cube, sphere, cylinder), color (8 values), material (rubber, metal), and size (small, large).

\subsection{CLEVR-Default in the Main Paper}

For the CLEVR-Default experiments in the main paper, we construct the concepts by generating 20 concepts organized as a chain-like procedure.
Here, each concept is derived from the previous one by changing the attributes of one or two objects, while keeping the overall scene layout highly similar across adjacent concepts.
In the benchmark used in the main paper, these edits are restricted to shape, color, and material changes, while object size and object positions are kept unchanged across neighboring concepts.
Each concept corresponds to a fixed object configuration and is rendered into 10 multi-view images, yielding 200 images in total.
For evaluation, we use one image per concept as the support example and the remaining nine as query images.

Table~\ref{tab:clevrv2_change_log} lists the actual scene-to-scene edits for one instantiation of this benchmark (denoted as CLEVR).
The table makes the chain construction explicit: each transition from scene $t$ to scene $t+1$ is realized by one or two local attribute edits on individual objects.
Most transitions involve a single modification, such as changing the color of object 2 from blue to green (scene00$\rightarrow$scene01) or the material of object 3 from metal to rubber (scene01$\rightarrow$scene02).
A smaller number of transitions involve two simultaneous edits, e.g., scene02$\rightarrow$scene03, scene08$\rightarrow$scene09, scene16$\rightarrow$scene17, and scene18$\rightarrow$scene19.
Importantly, all edits remain local and limited to shape, color, or material, so consecutive scenes stay highly similar despite representing different personalized concepts.

\begin{table*}[t]
\centering
\small
\setlength{\tabcolsep}{10pt} 

\caption{Scene-to-scene attribute change in CLEVR-Default. Each row indicates an actual edit applied when moving from scene $t$ to scene $t+1$.}
\label{tab:clevrv2_change_log}
\vspace{5pt}

\begin{tabular}{l l c l l l}
\hline
\textbf{From Scene} & \textbf{To Scene} & \textbf{Object idx} & \textbf{Attribute} & \textbf{From value} & \textbf{To value} \\
\hline
scene00 & scene01 & 2 & color    & blue      & green    \\
scene01 & scene02 & 3 & material & metal     & rubber   \\
scene02 & scene03 & 3 & shape    & cylinder & cube     \\
scene02 & scene03 & 4 & color    & gray     & yellow   \\
scene03 & scene04 & 0 & color    & red      & brown    \\
scene04 & scene05 & 2 & shape    & cube     & sphere   \\
scene05 & scene06 & 0 & material & rubber   & metal    \\
scene06 & scene07 & 1 & material & rubber   & metal    \\
scene07 & scene08 & 1 & color    & brown    & cyan     \\
scene08 & scene09 & 4 & material & rubber   & metal    \\
scene08 & scene09 & 0 & color    & brown    & purple   \\
scene09 & scene10 & 2 & color    & green    & blue     \\
scene10 & scene11 & 4 & color    & yellow   & red      \\
scene11 & scene12 & 3 & material & rubber   & metal    \\
scene12 & scene13 & 4 & shape    & cube     & sphere   \\
scene13 & scene14 & 1 & shape    & cylinder & sphere   \\
scene14 & scene15 & 0 & material & metal    & rubber   \\
scene15 & scene16 & 2 & shape    & sphere   & cylinder \\
scene16 & scene17 & 4 & color    & red      & gray     \\
scene16 & scene17 & 2 & material & rubber   & metal    \\
scene17 & scene18 & 3 & color    & red      & brown    \\
scene18 & scene19 & 4 & material & metal    & rubber   \\
scene18 & scene19 & 3 & material & metal    & rubber   \\
\hline
\vspace{-20pt}
\end{tabular}
\end{table*}

\subsection{Difficulty-controlled CLEVR}

In addition to the chain-based benchmark used in the main paper, we further construct \textbf{difficulty-controlled} CLEVR variants to analyze the behavior of personalization methods under systematically varying scene complexity and inter-concept distance.
These additional benchmarks are parameterized by the base number of objects $K \in \{4,5,6\}$ and by the hop size, which determines how far each derived scene is from its parent scene.

We define an ``edit unit'' as either changing one attribute (shape, color, size, or material) of a single object, or adding/removing one object.
A \textit{1-hop} modification applies one such edit unit, whereas a \textit{2-hop} modification applies two edit units sequentially, possibly on the same object or on different objects.
Using this definition, we generate two families of concept sets.

\begin{figure}[t]
    \centering
    \begin{minipage}{0.48\textwidth}
        \centering
        \includegraphics[width=\textwidth]{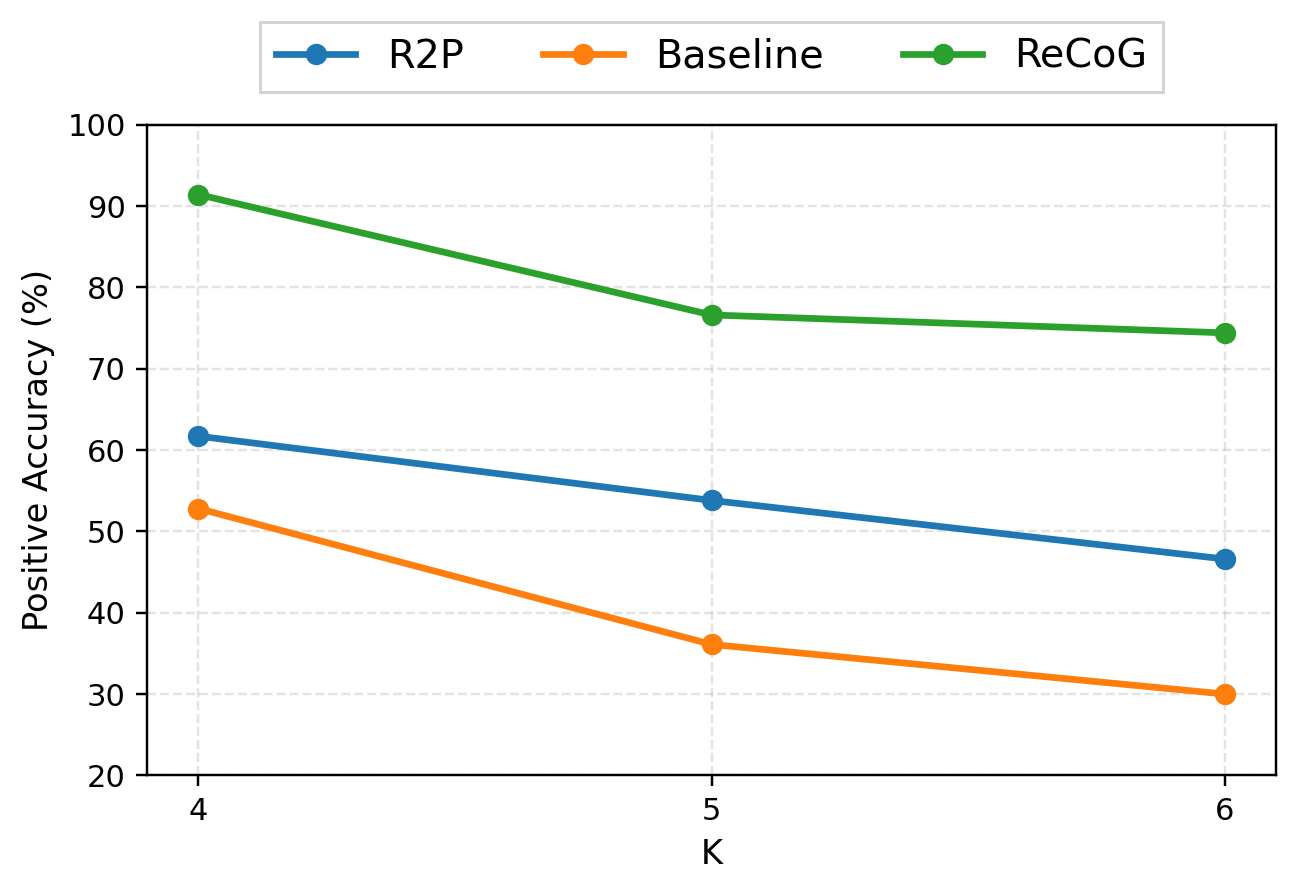}
        \vspace{-10pt}
        \caption{Performance comparison on difficulty-controlled CLEVR with different base scene complexities $K$ (1-hop).}
        \label{fig:1hop_line}
    \end{minipage}
    \hfill
    \begin{minipage}{0.48\textwidth}
        \centering
        \includegraphics[width=\textwidth]{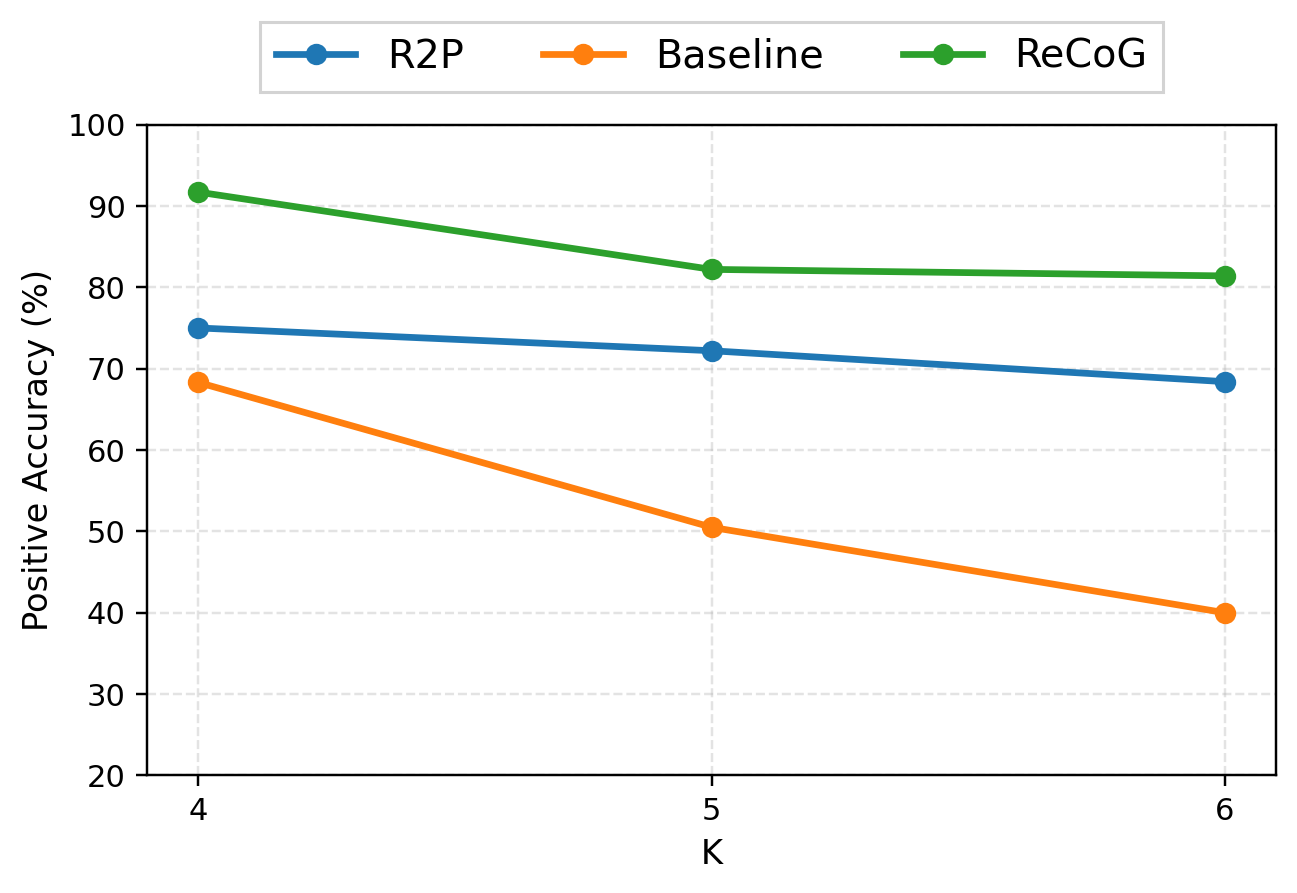}
        \vspace{-10pt}
        \caption{Performance comparison on difficulty-controlled CLEVR with different base scene complexities $K$ (2-hop).}
        \label{fig:2hop_line}
    \end{minipage}
\end{figure}

For the \textbf{1-hop} setting, we first sample a random root scene with $K$ objects.
We then generate four scenes by applying one 1-hop modification to the root.
From each of these four scenes, we further generate four additional scenes using another 1-hop modification.
After discarding the initial root scene, this yields a set of 20 scenes in total.

For the \textbf{2-hop} setting, we follow the same procedure, except that each expansion step uses a 2-hop modification instead of a 1-hop one.
Again, after removing the root scene, we obtain 20 scenes.
During generation, we explicitly remove duplicate scenes to avoid degenerate cycles or accidental returns to previously generated configurations.

Compared with the CLEVR benchmark in the main paper, these additional variants provide two explicit difficulty knobs.
First, increasing $K$ makes the scene more complex by increasing the number of objects.
Second, the hop size controls the separation between concepts: 1-hop concepts are more locally confusable, while 2-hop concepts are typically more separated in concept space.
This allows a more fine-grained analysis of how personalization methods behave as the task becomes either more cluttered or more subtle.


\subsection{Experimental Results on Difficulty-controlled CLEVR}

Table~\ref{tab:hop_main} in the main paper and Fig.~\ref{fig:1hop_line}-\ref{fig:2hop_line} summarize the results.
Several trends are consistent across all settings.
First, increasing $K$ consistently makes the task more difficult.
As the number of objects grows from 4 to 6, positive accuracy drops for all methods in both the 1-hop and 2-hop settings.
This confirms that larger scenes introduce more distractors and make fine-grained matching substantially harder.

Second, the 1-hop setting is more challenging than the 2-hop setting, especially in terms of positive accuracy.
Although 2-hop involves more edits, the resulting concepts are generally farther apart and therefore less confusable than 1-hop concepts, which differ by only a single local change.
This effect is clearly visible for both R2P and the baseline, whose positive accuracy improves noticeably in the 2-hop setting.

Third, ReCoG consistently achieves the best performance across all values of $K$ and under both hop settings.
The advantage is particularly large in the more confusable 1-hop regime.
For example, at $K=6$, ReCoG achieves 74.4\% positive accuracy in the 1-hop setting, compared with 46.6\% for R2P and 30.0\% for the baseline.
Even in the 2-hop setting, where concepts are relatively easier to separate, ReCoG still maintains a clear margin, reaching 81.4\% at $K=6$, compared with 68.4\% and 40.0\% for R2P and the baseline, respectively.

Finally, ReCoG shows notably stronger robustness as scene complexity increases from $K=5$ to $K=6$.
In both the 1-hop and 2-hop settings, its positive accuracy drops only marginally, whereas the competing methods suffer substantially larger degradations.
For example, in the 1-hop setting ReCoG decreases from 76.6\% to 74.4\%, while R2P drops from 53.8\% to 46.6\% and the baseline from 36.1\% to 30.0\%.
Likewise, in the 2-hop setting ReCoG remains nearly unchanged (82.2\% to 81.4\%), in contrast to the larger declines of R2P (72.2\% to 68.4\%) and the baseline (50.5\% to 40.0\%).
This suggests that ReCoG scales more gracefully to more cluttered scenes and preserves its discriminative ability even in harder personalization regimes.

\subsection{CLEVR with Changing Positional Arrangement}

We additionally consider a harder CLEVR setting in which the main source of variation is positional arrangement.
Concretely, we begin with the four scenes generated by a 1-hop modification and, for each of them, create four additional 1-hop variants.
In this second stage, the modification is restricted to changes in the \emph{xy positions} of objects, rather than their attributes or identities.
This setup makes the task significantly more difficult, because candidate scenes remain almost identical in object composition and differ mainly in spatial layout.

As expected, Table~\ref{tab:position} shows that the overall performance degrades under this setting, indicating that positional rearrangement is a particularly challenging factor for training-free personalization.
Nevertheless, ReCoG still consistently and substantially outperforms both the baseline and R2P, showing that explicit comparison remains effective even in this harder regime.
These results suggest that, while purely spatial changes without strong attribute differences are still difficult to resolve reliably, our relation-aware retrieval strategy provides a clear advantage over relation-agnostic alternatives.

\begin{table}[t]
\centering
\setlength{\tabcolsep}{3pt}
\caption{Performance comparison for experiments on changing positional arrangement. Values are reported in Positive Accuracy/Negative Accuracy (\%).}
\label{tab:position}
\vspace{10pt}
\small
\begin{tabular}{c|l|c}
\hline
\textbf{K} & \textbf{Methods} & P.Acc / N.Acc\\
\hline
\multirow{3}{*}{\textbf{4}} & R2P & 27.8 / 96.2 \\
 & Baseline & 18.0 / 95.7 \\
 & ReCoG & \textbf{33.2} / \textbf{96.5} \\
\hline
\end{tabular}
\end{table}

\begin{table}[t]
\centering
\caption{Comparative results on CUB-Hard: MiniCPM-o-2.6 vs. LLaVA-1.5. Bold indicates the best performance within each model group.}
\label{tab:comparison_llava_minicpm}
\vspace{10pt}
\small
\begin{tabular}{l|ccc|ccc}
\hline
\multirow{2}{*}{\textbf{Methods}} & \multicolumn{3}{c|}{\textbf{MiniCPM-o-2.6}} & \multicolumn{3}{c}{\textbf{LLaVA-1.5}} \\
\cline{2-7}
 & P.Acc & N.Acc & Wtd. & P.Acc & N.Acc & Wtd. \\
\hline
Baseline & 41.8 & 96.9 & 69.4 & 36.1 & 96.6 & 66.4 \\
\hline
ReCoG(Ours) & \textbf{86.3} & \textbf{99.3} & \textbf{92.8} & \textbf{89.4} & \textbf{99.5} & \textbf{94.5} \\
\hline
\end{tabular}
\end{table}

\section{Construction and Variance of CUB-Hard}
\label{sec:app_cub_construction_variance}

CUB contains 200 birds and 312 class-level visual attributes. Among these attributes, we identified 16 attributes corresponding to black coloration in different body regions, such as wing or upperparts. For each species, we computed the mean score across these 16 attributes and ranked all 200 species. CUB-Hard consists of the 20 species with the highest scores. The support/query images were randomly selected.

To analyze the variance of CUB-Hard, we constructed three alternative concept sets using the same procedure. Specifically, we selected white, brown, and grey coloration, respectively. ReCoG consistently outperforms R2P with low variation across concept sets (see Table~\ref{tab:cub_concept_variance}).

\begin{table}[h]
\centering
\begin{tabular}{l|cc}
\hline
\textbf{Concept set} & \textbf{R2P P.Acc.} & \textbf{ReCoG P.Acc.} \\
\hline
Black & 63.8\% & \textbf{86.3\%} \\
White & 64.5\% & \textbf{90.8\%} \\
Brown & 69.0\% & \textbf{95.3\%} \\
Grey  & 63.2\% & \textbf{92.1\%} \\
\hline
Mean $\pm$ std. & 65.1 $\pm$ 2.3\% & \textbf{91.1 $\pm$ 3.2\%} \\
\hline
\end{tabular}
\caption{Performance comparison across different concept sets based on coloration.}
\label{tab:cub_concept_variance}
\end{table}

\section{Impact of Employed VLMs}
\label{sec:app_impact_vlm}

To examine whether the effectiveness of ReCoG depends on a specific VLM backbone, we additionally replace the recognition model used in the main paper (MiniCPM-o-2.6) with LLaVA-1.5 and evaluate on CUB-Hard.
Table~\ref{tab:comparison_llava_minicpm} shows that ReCoG consistently outperforms the baseline under both backbones.
With MiniCPM-o-2.6, ReCoG improves the weighted accuracy from 69.4 to 92.8.
The same trend is preserved with LLaVA-1.5, where ReCoG improves the weighted accuracy from 66.4 to 94.5.
A similar pattern is observed for positive accuracy: under LLaVA-1.5, ReCoG achieves 89.4, substantially higher than the baseline score of 36.1.
These results indicate that the gain from ReCoG does not rely on a particular VLM choice.
Even when the recognition backbone is replaced, incorporating relation-aware concept retrieval consistently provides large improvements over the relation-agnostic baseline.
This suggests that ReCoG is a general and backbone-compatible personalization framework.

\section{Details on VLM Prompting}
\label{sec:app_prompt}

\subsection{Prompt Template for Relation-aware Database Construction}

In this section, we provide the prompt used for database construction ($P_\mathrm{disc}$ in Sec.~4.2). 
Figure~\ref{fig:prompt_db} shows the template for eliciting discriminative concept features from a frozen VLM.

\begin{figure}[t]
    \centering
    \begin{tcolorbox}[
        colback=gray!25!white, 
        colframe=gray!25!white, 
        arc=5pt, 
        left=15pt, right=15pt, 
        top=15pt, bottom=15pt, 
        boxrule=0pt
    ]
        \small \noindent
        Compare the two \{category\} concepts identified by the concept-identifiers and highlight what makes each concept unique. \\

        - Image 1: $<$example1$>$ \\
        - Image 2: $<$example2$>$ \\
        
        \vspace{5pt}
        \noindent
        Your response MUST follow EXACTLY the JSON format shown below (and nothing else): \\
        \texttt{\{} \\
        \texttt{\hspace*{1em}"example1 distinct features": ["feat1", "feat2", "feat3"],} \\
        \texttt{\hspace*{1em}"example2 distinct features": ["feat1", "feat2", "feat3"]} \\
        \texttt{\}} \\
        
        \vspace{5pt}
        \noindent
        \textbf{IMPORTANT:} \\
        - Do not include any additional text or commentary. \\
        - Any deviation from this format will be considered incorrect.
    \end{tcolorbox}
    \vspace{-10pt}
    \caption{Prompt template for relation-aware database construction.}
    \label{fig:prompt_db}
\end{figure}

Given two support images corresponding to a pair of concepts, the prompt asks the VLM to explicitly compare them and identify what makes each concept distinct from the other.
Unlike generic captioning or independent concept description, this prompt is designed to produce \emph{relational} and \emph{contrastive} evidence, i.e., features that are useful specifically for distinguishing one concept from a visually similar alternative.
This pairwise comparison is the key ingredient for constructing the directed edges in our relational concept graph.

To ensure that the VLM output can be parsed and stored reliably, we enforce a strict JSON-only response format.
For each ordered concept pair, the model returns two lists of distinct features, one for each concept from its own perspective.
These textual outputs are then used as the discriminative edge evidence.
By repeating this process over concept pairs, we obtain a database that captures not only per-concept visual representations, but also explicit inter-concept differences that are directly useful for fine-grained personalized recognition.

\subsection{Prompt Template for Comparison-based Retrieval}

\begin{figure}[t]
    \centering
    \begin{tcolorbox}[
        colback=gray!25!white, 
        colframe=gray!25!white, 
        arc=3pt, 
        left=10pt, right=10pt, 
        top=8pt, bottom=8pt, 
        boxrule=0pt
    ]
        \footnotesize \noindent
        You are a helpful AI agent specializing in image analysis and object recognition. \\
        You are given three images: \\
        - \textbf{Image 1:} Query image (the image to classify) \\
        - \textbf{Image 2:} Reference image for concept ``\{concept1\_name\}'' \\
        - \textbf{Image 3:} Reference image for concept ``\{concept2\_name\}'' \\
        
        \smallskip
        \noindent
        Additionally, you are provided with information about these two concepts: \\
        Distinctive features of \{concept1\_name\}: ["feat1", "feat2", "feat3"] \\
        Distinctive features of \{concept2\_name\}: ["feat1", "feat2", "feat3"] \\
        
        \vspace{5pt}
        \noindent
        \textbf{Task:} \\
        - Compare Image 1 (query) with Image 2 and Image 3. \\
        - Determine which concept the query image belongs to. \\
        - Focus on the distinctive features provided above to make your decision. \\
        
        \vspace{5pt}
        \noindent
        \textbf{Question:} \\
        - Provide your reasoning based on the distinctive features. \\
        - Generate your response in JSON format: \\ 
        \indent \qquad  \texttt{\{"Reasoning": "...", "Answer": "..."\}} \\
        - Output only the JSON response. DO NOT output any additional text.
    \end{tcolorbox}
    \vspace{-10pt}
    \caption{Prompt template for comparison-based retrieval.}
    \label{fig:prompt_structure_final}
\end{figure}

In this section, we provide the prompt used for discriminative retrieval via explicit comparison ($P_\mathrm{comp}$ in Sec.~4.3). Figure~\ref{fig:prompt_structure_final} shows the template used at inference time for comparison-based retrieval.

Given a query image and two candidate support images, the prompt asks a frozen VLM to determine which candidate concept is more compatible with the query.
Importantly, the model is not asked to make this decision based only on global visual similarity.
Instead, it is additionally provided with the concept-specific distinctive features extracted during database construction, so that the comparison can focus on the most discriminative evidence between the two candidates.
This design directly reflects the retrieval procedure of ReCoG.
After CLIP-based pruning produces a small candidate set, fine-grained ambiguity remains among visually similar concepts.
This prompt resolves such ambiguity by forcing the VLM to compare two candidates side by side while conditioning on their relation-aware textual evidence.
In this way, retrieval is performed through relative judgment rather than independent scoring of each concept.

For each comparison, the model returns a short reasoning trace together with its final choice. These pairwise decisions are then aggregated across all candidate pairs, as described in Sec.~4.3, to determine the final prediction.

\section{Validation on Realistic Open-Domain Personalized Concepts}
\label{sec:app_impact_vlm}

To further validate ReCoG, we additionally conducted an experiment on CelebA. We treated each identity ID as a personalized concept. For each concept, we used one support image and 19 query images. We constructed all 190 pairwise relational edges among the 20 concepts. Under support-image seed 42, ReCoG improves P.Acc. from 76.1\% to 94.2\% (see Table~\ref{tab:celeba_results}). This result provides additional evidence that ReCoG is effective not only for birds or synthetic scenes, but also for fine-grained recognition of real human identities.

\begin{table}[h]
\centering
\begin{tabular}{l|c|c|c}
\hline
Dataset & R2P P.Acc. & ReCoG P.Acc. & $\Delta$ \\
\hline
CelebA  & 76.1\%     & \textbf{94.2\%} & \textbf{+18.1} \\
\hline
\end{tabular}
\caption{Personalized recognition performance on CelebA.}
\label{tab:celeba_results}
\end{table}

\section{Sensitivity to the Selected Support Image}
\label{sec:app_support_sensitivity}

The method may be sensitive to the chosen support image. Since relational edges are generated from a single support image per concept, the extracted cues may reflect image-specific artifacts rather than stable concept-level differences. To address this issue, we resampled the support image for every concept using five different random seeds and reconstructed the corresponding concept databases. All other evaluation settings (CUB-Hard) were kept unchanged. ReCoG outperforms R2P under every support-image selection (see Table~\ref{tab:seed_sensitivity}). Although the absolute performance of ReCoG varies across exemplars, its advantage remains consistent. Moreover, we find that the performance variance of ReCoG is even smaller than that of R2P. These results demonstrate that the gain is not tied to one particular set of support images.

\begin{table}[h]
\centering
\begin{tabular}{l c c}
\hline
\textbf{Seed} & \textbf{R2P (P.Acc.)} & \textbf{ReCoG (P.Acc.)} \\
\hline
23            & 59.7\%                & 84.2\%                 \\
42            & 65.8\%                & 83.7\%                 \\
123           & 61.8\%                & 83.2\%                 \\
2024          & 72.4\%                & 90.8\%                 \\
2025          & 69.2\%                & 87.6\%                 \\
\hline
Mean$\pm$std. & 65.8$\pm$5.2\%        & 85.9$\pm$3.2\%         \\
\hline
\end{tabular}
\caption{Sensitivity to support image selection under different random seeds.}
\label{tab:seed_sensitivity}
\end{table}

\section{Qualitative Results}
\label{sec:app_qual}

\begin{figure}[t]
\centering
\includegraphics[width=\linewidth]{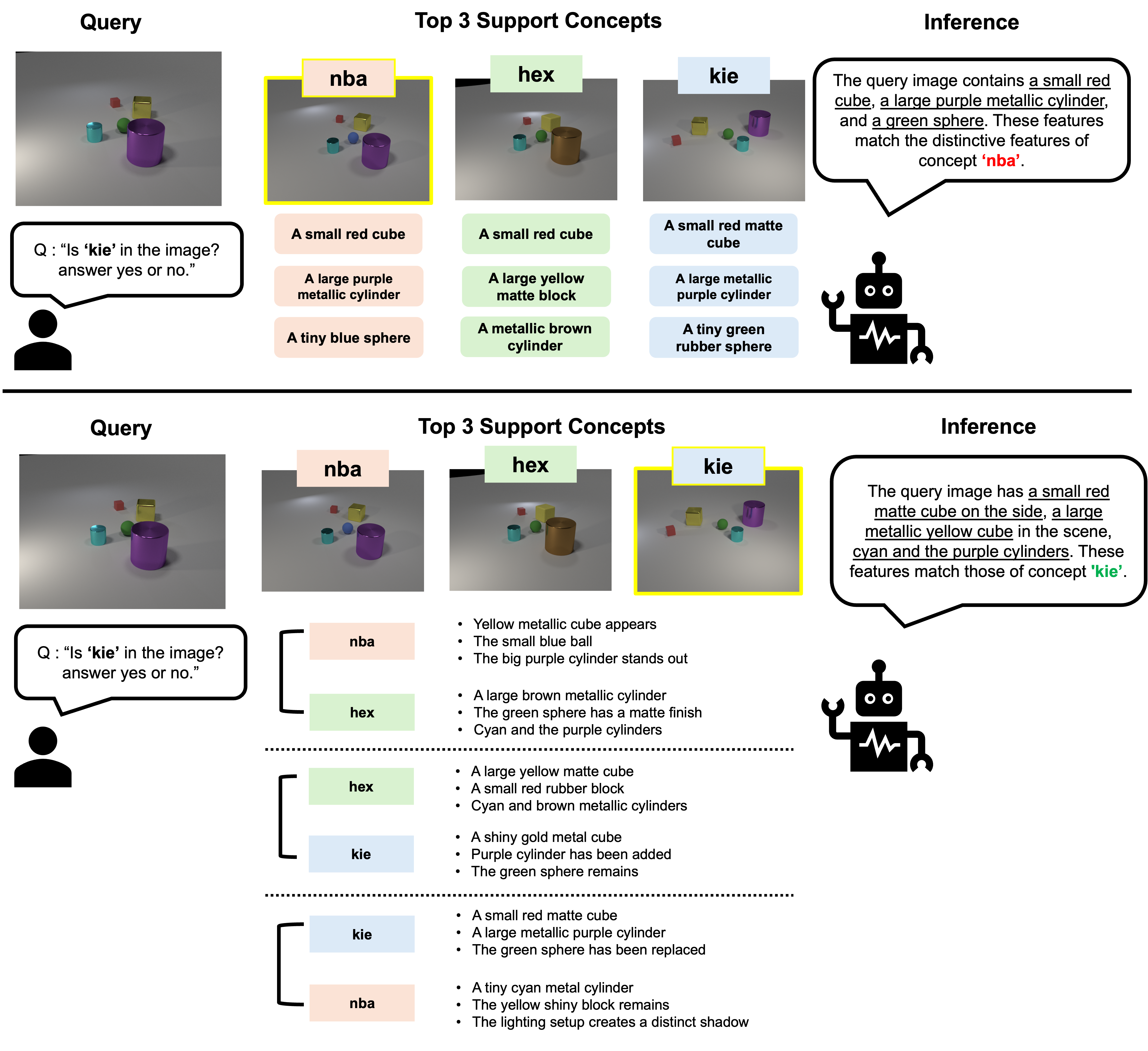}
\caption{Qualitative example comparing R2P-style retrieval (top) and ReCoG (bottom). The query belongs to \textit{kie}. While the relation-agnostic method is misled by prominent shared attributes and incorrectly favors \textit{nba}, ReCoG resolves the ambiguity through explicit pairwise comparison with relation-aware evidence and correctly predicts \textit{kie}.}
\label{fig:qual_supp}
\end{figure}

Figure~\ref{fig:qual_supp} in the next page illustrates why explicit comparison is important in fine-grained personalization. For reference, the answer is \texttt{Yes}.

In the relation-agnostic pipeline (top), the model reasons from independently described concept features and therefore tends to focus on salient attributes that are visible in the query, such as the small red cube, and the large purple metallic cylinder.
However, many of these cues are not truly discriminative because they are also compatible with competing concepts (\textit{hex} and \textit{nba}).
As a result, the prediction is driven by generic overlap rather than by the critical differences among candidates, ultimately leading to the wrong answer: \textit{nba}.

In contrast, ReCoG (bottom) explicitly compares candidates using relation-aware evidence stored between concept pairs.
This makes the model attend to how one candidate differs from another, rather than simply checking whether a candidate has some visible attributes.
For example, the comparison highlights subtle yet distinguishing cues such as "The green sphere has been replaced" (vs. \textit{nba}) or "Purple cylinder has been added" (vs. \textit{hex}).
These separate \textit{kie} from the other retrieved concepts.
By grounding the decision on these inter-concept differences, ReCoG correctly resolves the ambiguity and predicts \textit{kie}.

\begin{figure}[t]
\centering
\includegraphics[width=\linewidth]{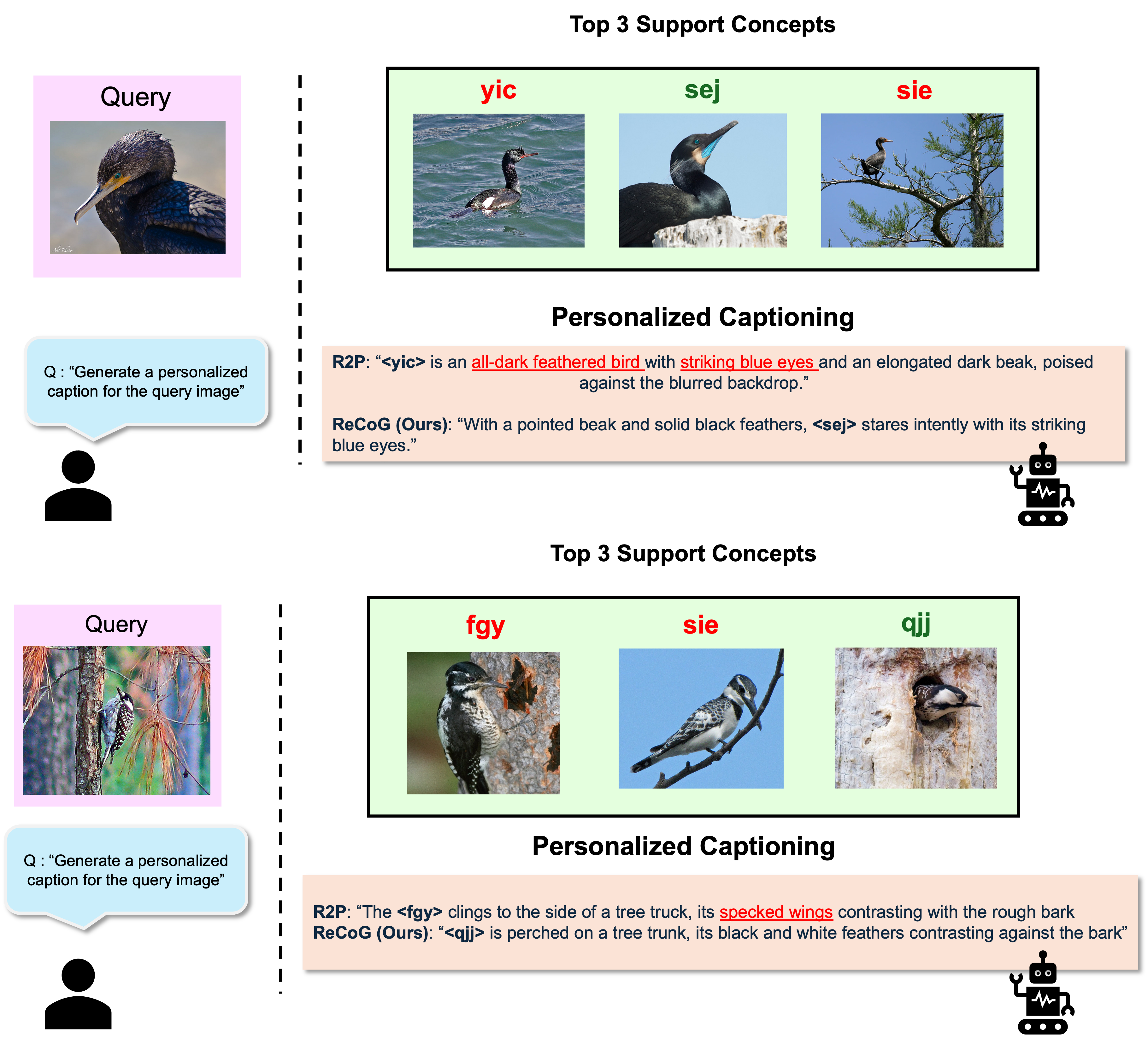}
\caption{Qualitative examples comparing R2P-style captioning and ReCoG.}
\label{fig:caption_results}
\end{figure}

In addition, Fig.~\ref{fig:caption_results} compares the captioning results of R2P and ReCoG.

\clearpage

\section*{NeurIPS Paper Checklist}

\begin{enumerate}

\item {\bf Claims}
    \item[] Question: Do the main claims made in the abstract and introduction accurately reflect the paper's contributions and scope?
    \item[] Answer: \answerYes{} 
    \item[] Justification: We present the contributions and scope of our work in abstract and introduction.
    \item[] Guidelines:
    \begin{itemize}
        \item The answer \answerNA{} means that the abstract and introduction do not include the claims made in the paper.
        \item The abstract and/or introduction should clearly state the claims made, including the contributions made in the paper and important assumptions and limitations. A \answerNo{} or \answerNA{} answer to this question will not be perceived well by the reviewers. 
        \item The claims made should match theoretical and experimental results, and reflect how much the results can be expected to generalize to other settings. 
        \item It is fine to include aspirational goals as motivation as long as it is clear that these goals are not attained by the paper. 
    \end{itemize}

\item {\bf Limitations}
    \item[] Question: Does the paper discuss the limitations of the work performed by the authors?
    \item[] Answer: \answerYes{} 
    \item[] Justification:  We discuss the limitations at the end of conclusion section.
    \item[] Guidelines:
    \begin{itemize}
        \item The answer \answerNA{} means that the paper has no limitation while the answer \answerNo{} means that the paper has limitations, but those are not discussed in the paper. 
        \item The authors are encouraged to create a separate ``Limitations'' section in their paper.
        \item The paper should point out any strong assumptions and how robust the results are to violations of these assumptions (e.g., independence assumptions, noiseless settings, model well-specification, asymptotic approximations only holding locally). The authors should reflect on how these assumptions might be violated in practice and what the implications would be.
        \item The authors should reflect on the scope of the claims made, e.g., if the approach was only tested on a few datasets or with a few runs. In general, empirical results often depend on implicit assumptions, which should be articulated.
        \item The authors should reflect on the factors that influence the performance of the approach. For example, a facial recognition algorithm may perform poorly when image resolution is low or images are taken in low lighting. Or a speech-to-text system might not be used reliably to provide closed captions for online lectures because it fails to handle technical jargon.
        \item The authors should discuss the computational efficiency of the proposed algorithms and how they scale with dataset size.
        \item If applicable, the authors should discuss possible limitations of their approach to address problems of privacy and fairness.
        \item While the authors might fear that complete honesty about limitations might be used by reviewers as grounds for rejection, a worse outcome might be that reviewers discover limitations that aren't acknowledged in the paper. The authors should use their best judgment and recognize that individual actions in favor of transparency play an important role in developing norms that preserve the integrity of the community. Reviewers will be specifically instructed to not penalize honesty concerning limitations.
    \end{itemize}

\item {\bf Theory assumptions and proofs}
    \item[] Question: For each theoretical result, does the paper provide the full set of assumptions and a complete (and correct) proof?
    \item[] Answer: \answerNA{} 
    \item[] Justification: Our work does not include theoretical results.
    \item[] Guidelines:
    \begin{itemize}
        \item The answer \answerNA{} means that the paper does not include theoretical results. 
        \item All the theorems, formulas, and proofs in the paper should be numbered and cross-referenced.
        \item All assumptions should be clearly stated or referenced in the statement of any theorems.
        \item The proofs can either appear in the main paper or the supplemental material, but if they appear in the supplemental material, the authors are encouraged to provide a short proof sketch to provide intuition. 
        \item Inversely, any informal proof provided in the core of the paper should be complemented by formal proofs provided in appendix or supplemental material.
        \item Theorems and Lemmas that the proof relies upon should be properly referenced. 
    \end{itemize}

    \item {\bf Experimental result reproducibility}
    \item[] Question: Does the paper fully disclose all the information needed to reproduce the main experimental results of the paper to the extent that it affects the main claims and/or conclusions of the paper (regardless of whether the code and data are provided or not)?
    \item[] Answer: \answerYes{} 
    \item[] Justification: We present all experimental settings and the information necessary to reproduce the main results.
    \item[] Guidelines:
    \begin{itemize}
        \item The answer \answerNA{} means that the paper does not include experiments.
        \item If the paper includes experiments, a \answerNo{} answer to this question will not be perceived well by the reviewers: Making the paper reproducible is important, regardless of whether the code and data are provided or not.
        \item If the contribution is a dataset and\slash or model, the authors should describe the steps taken to make their results reproducible or verifiable. 
        \item Depending on the contribution, reproducibility can be accomplished in various ways. For example, if the contribution is a novel architecture, describing the architecture fully might suffice, or if the contribution is a specific model and empirical evaluation, it may be necessary to either make it possible for others to replicate the model with the same dataset, or provide access to the model. In general. releasing code and data is often one good way to accomplish this, but reproducibility can also be provided via detailed instructions for how to replicate the results, access to a hosted model (e.g., in the case of a large language model), releasing of a model checkpoint, or other means that are appropriate to the research performed.
        \item While NeurIPS does not require releasing code, the conference does require all submissions to provide some reasonable avenue for reproducibility, which may depend on the nature of the contribution. For example
        \begin{enumerate}
            \item If the contribution is primarily a new algorithm, the paper should make it clear how to reproduce that algorithm.
            \item If the contribution is primarily a new model architecture, the paper should describe the architecture clearly and fully.
            \item If the contribution is a new model (e.g., a large language model), then there should either be a way to access this model for reproducing the results or a way to reproduce the model (e.g., with an open-source dataset or instructions for how to construct the dataset).
            \item We recognize that reproducibility may be tricky in some cases, in which case authors are welcome to describe the particular way they provide for reproducibility. In the case of closed-source models, it may be that access to the model is limited in some way (e.g., to registered users), but it should be possible for other researchers to have some path to reproducing or verifying the results.
        \end{enumerate}
    \end{itemize}

\item {\bf Open access to data and code}
    \item[] Question: Does the paper provide open access to the data and code, with sufficient instructions to faithfully reproduce the main experimental results, as described in supplemental material?
    \item[] Answer: \answerYes{} 
    \item[] Justification: The data and code will be publicly available upon acceptance.
    \item[] Guidelines:
    \begin{itemize}
        \item The answer \answerNA{} means that paper does not include experiments requiring code.
        \item Please see the NeurIPS code and data submission guidelines (\url{https://neurips.cc/public/guides/CodeSubmissionPolicy}) for more details.
        \item While we encourage the release of code and data, we understand that this might not be possible, so \answerNo{} is an acceptable answer. Papers cannot be rejected simply for not including code, unless this is central to the contribution (e.g., for a new open-source benchmark).
        \item The instructions should contain the exact command and environment needed to run to reproduce the results. See the NeurIPS code and data submission guidelines (\url{https://neurips.cc/public/guides/CodeSubmissionPolicy}) for more details.
        \item The authors should provide instructions on data access and preparation, including how to access the raw data, preprocessed data, intermediate data, and generated data, etc.
        \item The authors should provide scripts to reproduce all experimental results for the new proposed method and baselines. If only a subset of experiments are reproducible, they should state which ones are omitted from the script and why.
        \item At submission time, to preserve anonymity, the authors should release anonymized versions (if applicable).
        \item Providing as much information as possible in supplemental material (appended to the paper) is recommended, but including URLs to data and code is permitted.
    \end{itemize}

\item {\bf Experimental setting/details}
    \item[] Question: Does the paper specify all the training and test details (e.g., data splits, hyperparameters, how they were chosen, type of optimizer) necessary to understand the results?
    \item[] Answer: \answerYes{} 
    \item[] Justification: We specify all the details.
    \item[] Guidelines:
    \begin{itemize}
        \item The answer \answerNA{} means that the paper does not include experiments.
        \item The experimental setting should be presented in the core of the paper to a level of detail that is necessary to appreciate the results and make sense of them.
        \item The full details can be provided either with the code, in appendix, or as supplemental material.
    \end{itemize}

\item {\bf Experiment statistical significance}
    \item[] Question: Does the paper report error bars suitably and correctly defined or other appropriate information about the statistical significance of the experiments?
    \item[] Answer: \answerNA{} 
    \item[] Justification: As the proposed method is fully deterministic, reporting statistical significance is less necessary.
    \item[] Guidelines:
    \begin{itemize}
        \item The answer \answerNA{} means that the paper does not include experiments.
        \item The authors should answer \answerYes{} if the results are accompanied by error bars, confidence intervals, or statistical significance tests, at least for the experiments that support the main claims of the paper.
        \item The factors of variability that the error bars are capturing should be clearly stated (for example, train/test split, initialization, random drawing of some parameter, or overall run with given experimental conditions).
        \item The method for calculating the error bars should be explained (closed form formula, call to a library function, bootstrap, etc.)
        \item The assumptions made should be given (e.g., Normally distributed errors).
        \item It should be clear whether the error bar is the standard deviation or the standard error of the mean.
        \item It is OK to report 1-sigma error bars, but one should state it. The authors should preferably report a 2-sigma error bar than state that they have a 96\% CI, if the hypothesis of Normality of errors is not verified.
        \item For asymmetric distributions, the authors should be careful not to show in tables or figures symmetric error bars that would yield results that are out of range (e.g., negative error rates).
        \item If error bars are reported in tables or plots, the authors should explain in the text how they were calculated and reference the corresponding figures or tables in the text.
    \end{itemize}

\item {\bf Experiments compute resources}
    \item[] Question: For each experiment, does the paper provide sufficient information on the computer resources (type of compute workers, memory, time of execution) needed to reproduce the experiments?
    \item[] Answer: \answerYes{} 
    \item[] Justification: We provide the relevant information.
    \item[] Guidelines:
    \begin{itemize}
        \item The answer \answerNA{} means that the paper does not include experiments.
        \item The paper should indicate the type of compute workers CPU or GPU, internal cluster, or cloud provider, including relevant memory and storage.
        \item The paper should provide the amount of compute required for each of the individual experimental runs as well as estimate the total compute. 
        \item The paper should disclose whether the full research project required more compute than the experiments reported in the paper (e.g., preliminary or failed experiments that didn't make it into the paper). 
    \end{itemize}
    
\item {\bf Code of ethics}
    \item[] Question: Does the research conducted in the paper conform, in every respect, with the NeurIPS Code of Ethics \url{https://neurips.cc/public/EthicsGuidelines}?
    \item[] Answer: \answerYes{} 
    \item[] Justification: We respect the NeurIPS Code of Ethics.
    \item[] Guidelines:
    \begin{itemize}
        \item The answer \answerNA{} means that the authors have not reviewed the NeurIPS Code of Ethics.
        \item If the authors answer \answerNo, they should explain the special circumstances that require a deviation from the Code of Ethics.
        \item The authors should make sure to preserve anonymity (e.g., if there is a special consideration due to laws or regulations in their jurisdiction).
    \end{itemize}

\item {\bf Broader impacts}
    \item[] Question: Does the paper discuss both potential positive societal impacts and negative societal impacts of the work performed?
    \item[] Answer: \answerNA{} 
    \item[] Justification: We believe that this work does not present any notable societal impact.
    \item[] Guidelines:
    \begin{itemize}
        \item The answer \answerNA{} means that there is no societal impact of the work performed.
        \item If the authors answer \answerNA{} or \answerNo, they should explain why their work has no societal impact or why the paper does not address societal impact.
        \item Examples of negative societal impacts include potential malicious or unintended uses (e.g., disinformation, generating fake profiles, surveillance), fairness considerations (e.g., deployment of technologies that could make decisions that unfairly impact specific groups), privacy considerations, and security considerations.
        \item The conference expects that many papers will be foundational research and not tied to particular applications, let alone deployments. However, if there is a direct path to any negative applications, the authors should point it out. For example, it is legitimate to point out that an improvement in the quality of generative models could be used to generate Deepfakes for disinformation. On the other hand, it is not needed to point out that a generic algorithm for optimizing neural networks could enable people to train models that generate Deepfakes faster.
        \item The authors should consider possible harms that could arise when the technology is being used as intended and functioning correctly, harms that could arise when the technology is being used as intended but gives incorrect results, and harms following from (intentional or unintentional) misuse of the technology.
        \item If there are negative societal impacts, the authors could also discuss possible mitigation strategies (e.g., gated release of models, providing defenses in addition to attacks, mechanisms for monitoring misuse, mechanisms to monitor how a system learns from feedback over time, improving the efficiency and accessibility of ML).
    \end{itemize}
    
\item {\bf Safeguards}
    \item[] Question: Does the paper describe safeguards that have been put in place for responsible release of data or models that have a high risk for misuse (e.g., pre-trained language models, image generators, or scraped datasets)?
    \item[] Answer: \answerNA{} 
    \item[] Justification: Our paper poses no such risks.
    \item[] Guidelines:
    \begin{itemize}
        \item The answer \answerNA{} means that the paper poses no such risks.
        \item Released models that have a high risk for misuse or dual-use should be released with necessary safeguards to allow for controlled use of the model, for example by requiring that users adhere to usage guidelines or restrictions to access the model or implementing safety filters. 
        \item Datasets that have been scraped from the Internet could pose safety risks. The authors should describe how they avoided releasing unsafe images.
        \item We recognize that providing effective safeguards is challenging, and many papers do not require this, but we encourage authors to take this into account and make a best faith effort.
    \end{itemize}

\item {\bf Licenses for existing assets}
    \item[] Question: Are the creators or original owners of assets (e.g., code, data, models), used in the paper, properly credited and are the license and terms of use explicitly mentioned and properly respected?
    \item[] Answer: \answerYes{} 
    \item[] Justification: We explicitly mention them.
    \item[] Guidelines:
    \begin{itemize}
        \item The answer \answerNA{} means that the paper does not use existing assets.
        \item The authors should cite the original paper that produced the code package or dataset.
        \item The authors should state which version of the asset is used and, if possible, include a URL.
        \item The name of the license (e.g., CC-BY 4.0) should be included for each asset.
        \item For scraped data from a particular source (e.g., website), the copyright and terms of service of that source should be provided.
        \item If assets are released, the license, copyright information, and terms of use in the package should be provided. For popular datasets, \url{paperswithcode.com/datasets} has curated licenses for some datasets. Their licensing guide can help determine the license of a dataset.
        \item For existing datasets that are re-packaged, both the original license and the license of the derived asset (if it has changed) should be provided.
        \item If this information is not available online, the authors are encouraged to reach out to the asset's creators.
    \end{itemize}

\item {\bf New assets}
    \item[] Question: Are new assets introduced in the paper well documented and is the documentation provided alongside the assets?
    \item[] Answer: \answerYes{} 
    \item[] Justification: The newly proposed benchmark is explained in detail. Also, the documentation will be released upon acceptance.
    \item[] Guidelines:
    \begin{itemize}
        \item The answer \answerNA{} means that the paper does not release new assets.
        \item Researchers should communicate the details of the dataset\slash code\slash model as part of their submissions via structured templates. This includes details about training, license, limitations, etc. 
        \item The paper should discuss whether and how consent was obtained from people whose asset is used.
        \item At submission time, remember to anonymize your assets (if applicable). You can either create an anonymized URL or include an anonymized zip file.
    \end{itemize}

\item {\bf Crowdsourcing and research with human subjects}
    \item[] Question: For crowdsourcing experiments and research with human subjects, does the paper include the full text of instructions given to participants and screenshots, if applicable, as well as details about compensation (if any)? 
    \item[] Answer: \answerNA{} 
    \item[] Justification: Our work does not include such issue.
    \item[] Guidelines:
    \begin{itemize}
        \item The answer \answerNA{} means that the paper does not involve crowdsourcing nor research with human subjects.
        \item Including this information in the supplemental material is fine, but if the main contribution of the paper involves human subjects, then as much detail as possible should be included in the main paper. 
        \item According to the NeurIPS Code of Ethics, workers involved in data collection, curation, or other labor should be paid at least the minimum wage in the country of the data collector. 
    \end{itemize}

\item {\bf Institutional review board (IRB) approvals or equivalent for research with human subjects}
    \item[] Question: Does the paper describe potential risks incurred by study participants, whether such risks were disclosed to the subjects, and whether Institutional Review Board (IRB) approvals (or an equivalent approval/review based on the requirements of your country or institution) were obtained?
    \item[] Answer: \answerNA{} 
    \item[] Justification: Our work does not include such issue.
    \item[] Guidelines:
    \begin{itemize}
        \item The answer \answerNA{} means that the paper does not involve crowdsourcing nor research with human subjects.
        \item Depending on the country in which research is conducted, IRB approval (or equivalent) may be required for any human subjects research. If you obtained IRB approval, you should clearly state this in the paper. 
        \item We recognize that the procedures for this may vary significantly between institutions and locations, and we expect authors to adhere to the NeurIPS Code of Ethics and the guidelines for their institution. 
        \item For initial submissions, do not include any information that would break anonymity (if applicable), such as the institution conducting the review.
    \end{itemize}

\item {\bf Declaration of LLM usage}
    \item[] Question: Does the paper describe the usage of LLMs if it is an important, original, or non-standard component of the core methods in this research? Note that if the LLM is used only for writing, editing, or formatting purposes and does \emph{not} impact the core methodology, scientific rigor, or originality of the research, declaration is not required.
    \item[] Answer: \answerYes{} 
    \item[] Justification: We explicitly use VLM as a core component, and describe its usage in detail.
    \item[] Guidelines:
    \begin{itemize}
        \item The answer \answerNA{} means that the core method development in this research does not involve LLMs as any important, original, or non-standard components.
        \item Please refer to our LLM policy in the NeurIPS handbook for what should or should not be described.
    \end{itemize}

\end{enumerate}

\end{document}